\documentclass[letterpaper, 10 pt, conference]{ieeeconf}  % Comment this line out if you need a4paper

\IEEEoverridecommandlockouts                              % This command is only needed if 
\usepackage{graphics} % for pdf, bitmapped graphics files
\usepackage{epsfig} % for postscript graphics files
\usepackage{times} % assumes new font selection scheme installed
\usepackage{amsmath} % assumes amsmath package installed
\usepackage{amssymb}  % assumes amsmath package installed
\usepackage[backend=biber,sorting=nyt,sortcites=true]{biblatex}
\usepackage{multirow}
\usepackage{multicol}
\usepackage{hyperref}
\usepackage{booktabs}
\usepackage{enumerate}
\usepackage{color}
\usepackage{tabularx}
\usepackage{hyperref}
\usepackage{pifont}
\usepackage{algorithm}
\usepackage{algpseudocode}
\usepackage{amsmath}
\usepackage{adjustbox}
\usepackage{pgfplots}
\pgfplotsset{compat=1.18}
\usepackage[usenames,dvipsnames,table]{xcolor}
\hypersetup{
    colorlinks=true,
    linkcolor=MidnightBlue,
    filecolor=magenta,      
    urlcolor=MidnightBlue,
    citecolor=MidnightBlue,
} 
\usepackage[font=small,labelfont=bf]{caption}
\usepackage{makecell}
\usepackage{multirow}
\definecolor{highfreq}{RGB}{235,235,235}

\newcommand{\xmark}{\ding{55}}
\newcommand{\mate}{MATE}

\title{\LARGE \bf
MATE: Asynchronous Multi-Rate Diffusion Policy Composition 
}

\author{
Zihao He$^{*,1}$, Hongjie Fang$^{*,1}$, Shirun Tang$^2$, Cewu Lu$^{1,2,3,\dagger}$, Hao-Shu Fang$^{\triangle,\dagger}$
\thanks{$^*$Equal Contribution.\quad $^\dagger$Corresponding Authors.}
\thanks{$^1$Shanghai Jiao Tong University. \quad $^2$Noematrix.}
\thanks{$^3$Shanghai Innovation Institute. \quad $\triangle$Independent Researcher.}
}

\makeatletter

\let\oldmaketitle\@maketitle

\renewcommand{\@maketitle}{%
    \oldmaketitle

    \vspace{-0.4cm}

    \begin{center}
        \begin{minipage}{\textwidth}
            \centering

            \includegraphics[width=\linewidth]{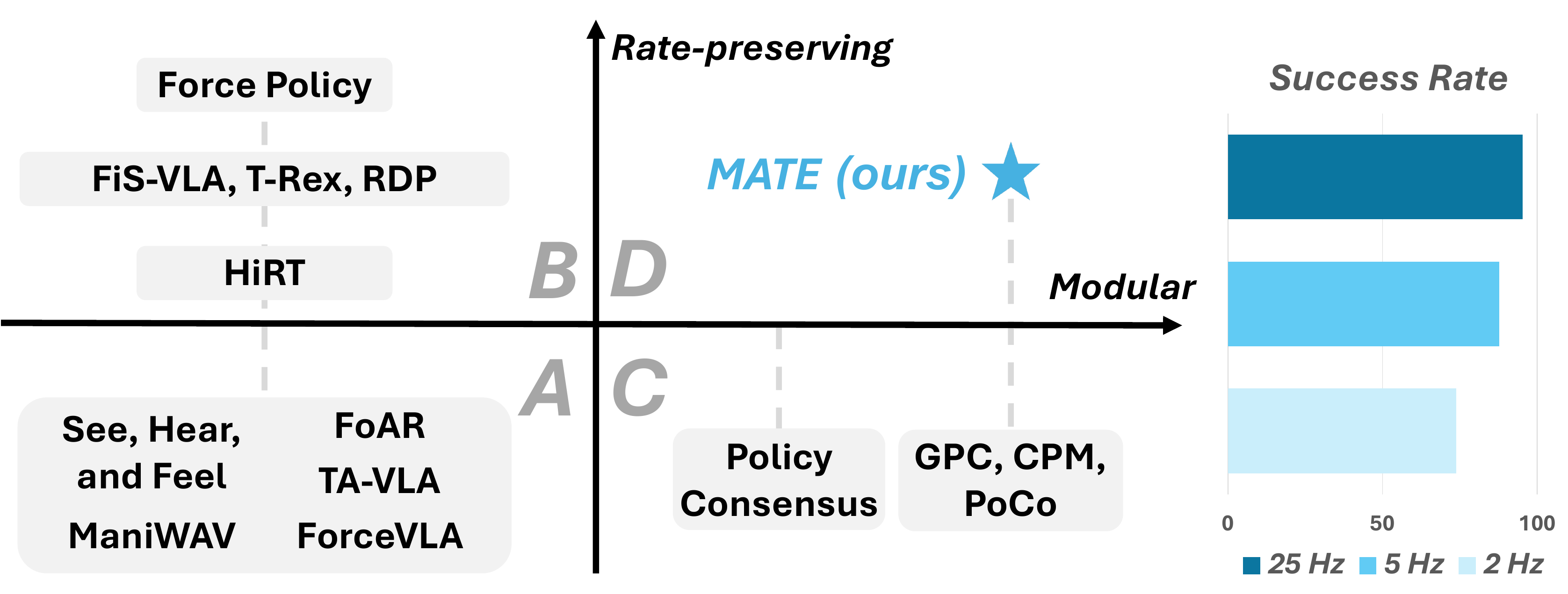}

            \vspace{-0.05cm}

            \captionof{figure}{\textbf{MATE enables modular, rate-preserving policy composition.}
            \textit{\textbf{(Left)}} We organize representative manipulation methods along two complementary properties: \textit{modularity} and \textit{rate preservation}.
            (A) Unified policies are neither modular nor rate-preserving.
            (B) Slow-fast and hierarchical methods preserve heterogeneous rates but rely on coupled architectures.
            (C) Modular composition methods combine independently trained components but synchronize them at a shared inference rate.
            (D) MATE achieves both properties through asynchronous multi-rate policy composition.
            \textit{\textbf{(Right)}} As a motivating experiment, we evaluate a standalone force policy on the \textit{Flip} task across five different objects, with the gripper initialized near the object in a pre-contact configuration.
            Higher policy inference rates consistently improve task success, motivating the need to preserve heterogeneous inference rates during policy composition.   
            }
            \label{fig:teaser}

        \end{minipage}
    \end{center}

    \vspace{-0.5cm}
}

\makeatother

\begin{document}

\maketitle
\setcounter{figure}{1} % Teaser is Fig. 1; the next figure is Fig. 2
\thispagestyle{empty}
\pagestyle{empty}

%%%%%%%%%%%%%%%%%%%%%%%%%%%%%%%%%%%%%%%%%%%%%%%%%%%%%%%%%%%%%%%%%%%%%%%%%%%%%%%%
\begin{abstract}
Diffusion policies have shown strong potential for robotic imitation learning, but different policy components can benefit from substantially different inference rates.
For example, visual policies provide rich global guidance but are computationally expensive, whereas policies conditioned on low-dimensional signals can be queried more frequently for reactive correction.
However, existing diffusion policies either synchronize all components to the slowest one or rely on jointly trained, task-specific multi-rate architectures that are difficult to extend.
We propose \mate{}, a modular framework for asynchronous multi-rate diffusion policy composition.
\mate{} allows independently trained component policies to operate at their native inference rates and contribute denoising guidance whenever new predictions become available.
To consistently compose predictions generated at different robot states, we derive a reference-frame rebasing rule for diffusion variables under relative action representations and align delayed guidance before fusion.
We instantiate \mate{} in contact-rich manipulation by composing a low-frequency vision policy with a high-frequency force policy, and further demonstrate its generality through asynchronous composition of complementary 3D and 2D visual policies.
Across multiple real-world tasks, \mate{} outperforms various baselines, demonstrating benefits beyond high-frequency inference alone.
\end{abstract}

%%%%%%%%%%%%%%%%%%%%%%%%%%%%%%%%%%%%%%%%%%%%%%%%%%%%%%%%%%%%%%%%%%%%%%%%%%%%%%%%

\section{Introduction}

Diffusion policies~\cite{diffusionpolicy, rise} have emerged as a powerful paradigm for robotic imitation learning, generating expressive action trajectories through iterative denoising~\cite{ddpm, ddim}.
Recent studies extend diffusion policies with additional sensory and task information, including point clouds, language instructions, force/torque measurements, tactile signals, and audio, to improve manipulation across diverse settings~\cite{foar, rdp, maniwav, crdagger, acp}.
However, existing work primarily focuses on \emph{what information} should be incorporated, while paying less attention to \emph{how frequently} different policy components should be queried during closed-loop execution.

These considerations motivate two desirable properties for combining policy components:
\begin{itemize}
    \item \textbf{Modular.}
    Modularity enables independently trained policy components to be reused and composed without joint retraining or redesigning the architecture.
    This supports incremental extension: new sensing capabilities or specialized behaviors can be incorporated by adding a complementary policy while retaining existing components.
    For example, an existing VLA policy could be augmented with a specialized force-conditioned policy for contact-aware correction \textit{without extra retraining}.

    \item \textbf{Rate-preserving.}
    Different policy components may naturally operate at different inference rates due to their sensing characteristics.
    Our motivated experiment in Fig.~\ref{fig:teaser} shows that higher policy inference frequency improves contact-stage success.
    This highlights the importance of preserving the native inference rate of each policy, allowing fast components to provide frequent corrections without being bottlenecked by slow ones.

\end{itemize}

% Different policy components can have substantially different temporal characteristics.
% Vision policies provide spatially rich, long-horizon guidance but often involve computationally expensive perception and encoding.
% In contrast, policies conditioned on low-dimensional signals, such as force feedback, are lightweight and can be queried much more frequently for local reactive correction.
% These components therefore need not, and often should not, operate at a single shared inference rate.

% \begin{figure}[t]
%     \centering
%     \includegraphics[width=\columnwidth]{images/lowdim_ft_policy_bar.pdf}
%     \caption{\textbf{Inference Frequency Motivation.}
%     We evaluate a standalone low-dimensional force-only policy on \textit{Flip} under different input and inference frequencies in a controlled contact-stage setting.
%     Increasing the inference frequency generally improves performance.}
%     \label{fig:force_frequency}
%     \vspace{-0.4cm}
% \end{figure}

Existing approaches do not fully satisfy both properties.
As summarized in Fig.~\ref{fig:teaser}, unified policies~\cite{tavla, foar, forcevla, maniwav, seehearfeel} couple multiple inputs within a shared inference pipeline, and do not explicitly preserve heterogeneous inference rates.
Slow-fast and hierarchical architectures~\cite{rdp, forcepolicy, hirt, trex, fastinslow} preserve heterogeneous update rates by introducing fast reactive components alongside slower ones.
However, these components are typically designed as coupled architectures, making it difficult to incorporate additional independently trained policies without architectural modification.

Policy composition offers a more modular alternative by combining independently trained policies at inference time~\cite{multimodal_consensus, poco, gpc, cpm}.
However, existing composition methods generally perform synchronous composition, where participating components are evaluated and combined at common inference steps.
They therefore do not explicitly preserve heterogeneous component rates, and the composed system is effectively constrained by the slower components.
As illustrated by the motivating experiment in Fig.~\ref{fig:teaser}, this can be undesirable when a lightweight policy benefits from more frequent inference.
This synchronization assumption breaks in a multi-rate setting: a slow policy prediction may be reused while a fast policy is queried multiple times, causing their action trajectories and denoising guidance to correspond to different timestamps and potentially different action reference frames.
Directly composing such predictions is geometrically inconsistent, while waiting for all policies would again reduce execution to the slowest inference rate.

To address this problem, we propose \mate{}, a framework for asynchronous multi-rate diffusion policy composition.
\mate{} decouples component-policy inference from action composition, allowing each policy to operate at its native rate and contribute denoising guidance whenever a new prediction becomes available.
To consistently compose guidance generated at different robot states, we derive a principled reference-frame rebasing rule for clean actions, noisy diffusion variables, and predicted noise under relative action representations.
The aligned guidance is then composed asynchronously, enabling high-frequency action updates without repeatedly querying slow policy components or requiring joint retraining.

We instantiate \mate{} in contact-rich manipulation by composing a low-frequency vision policy with a high-frequency force/torque policy.
The vision policy supplies global spatial and long-horizon task guidance, while the force policy provides rapid local corrections during physical interaction.
We further demonstrate that the framework is not restricted to vision-force fusion by composing complementary 3D and 2D visual policies in an occluded pick-and-place task.
Across the evaluated tasks, \mate{} outperforms synchronous composition, specially designed multimodal policies, and a matched-frequency end-to-end baseline.
Our main contributions are summarized as follows:
\begin{itemize}
    \item We formulate \emph{asynchronous multi-rate diffusion policy composition}, allowing independently trained policy components to operate at its native inference rates without being bottlenecked by the slowest component.
    \item We derive a principled reference-frame rebasing rule for diffusion variables and combine it with temporal overlap selection, enabling composition of denoising guidance generated at different query times and robot states.
    \item We demonstrate the framework through multi-rate vision-force and 3D-2D vision policy composition, and show improved performance over baselines.
\end{itemize}

\section{Related Works}

\subsection{Policy Composition in Robotics}
Policy composition has been widely explored as a way to combine multiple generative models, constraints, or guidance signals at inference time. Related ideas have appeared in visual generation~\cite{compositional_visual_generation, visualanagrams, compositional_ebm, geng2024factorized, images_that_sounds}, language modeling~\cite{dexperts, fudge, gedi, neurologic}, and planning~\cite{generative_skill_chaining, cdgs, ebm_zero_shot_planner}, where different semantic or structural conditions are composed to guide generation. In robotics, recent works extend this idea to policy learning. PoCo~\cite{poco} composes diffusion policies trained from heterogeneous robot datasets, while General Policy Composition~\cite{gpc} combines diffusion- or flow-based robot policies through test-time distribution-level composition. In addition, Policy Consensus~\cite{multimodal_consensus} learns router weights to adaptively combine modality-specific guidance at inference time.

However, existing robot policy composition methods are frequency-agnostic ~\cite{gpc, multimodal_consensus, poco, fdp}. They assume that all modalities are queried at the same time, in the same reference frame. This assumption simplifies composition, but it also implicitly requires all policies to operate at a shared decision frequency. As a result, the composed policy is often limited by the slowest modality or policy component.
Contact-rich manipulation is a representative example where this synchronization assumption becomes problematic. Stable physical interaction always requires fast local corrections from force/torque feedback.
MATE addresses this gap through asynchronous multi-rate composition, allowing independently trained policies to contribute guidance at their native inference rates while aligning their predictions in time and action reference frame.

% However, policy composition for contact-rich manipulation remains less explored. Existing methods typically compose policies at the same decision time, action horizon, and reference frame, which makes the overall control frequency limited by the slowest policy, usually vision-based models~\cite{poco, gpc, fdp, multimodal_consensus}. This is problematic for contact-rich manipulation, where force/torque feedback can be processed at much higher frequency and fast local reactions are critical for stable physical interaction~\cite{rdp, tacdiffusion, foar}. Our work addresses this limitation with Latency-Aware Guidance Fusion allowing asynchronous inference and fuse their diffusion variables only over the overlapping horizon. 
% Our method lets the slow vision policy provide global guidance while the fast force/torque policy supplies high-frequency local corrections.

\begin{figure*}[t]
    \centering
    \includegraphics[width=\textwidth]{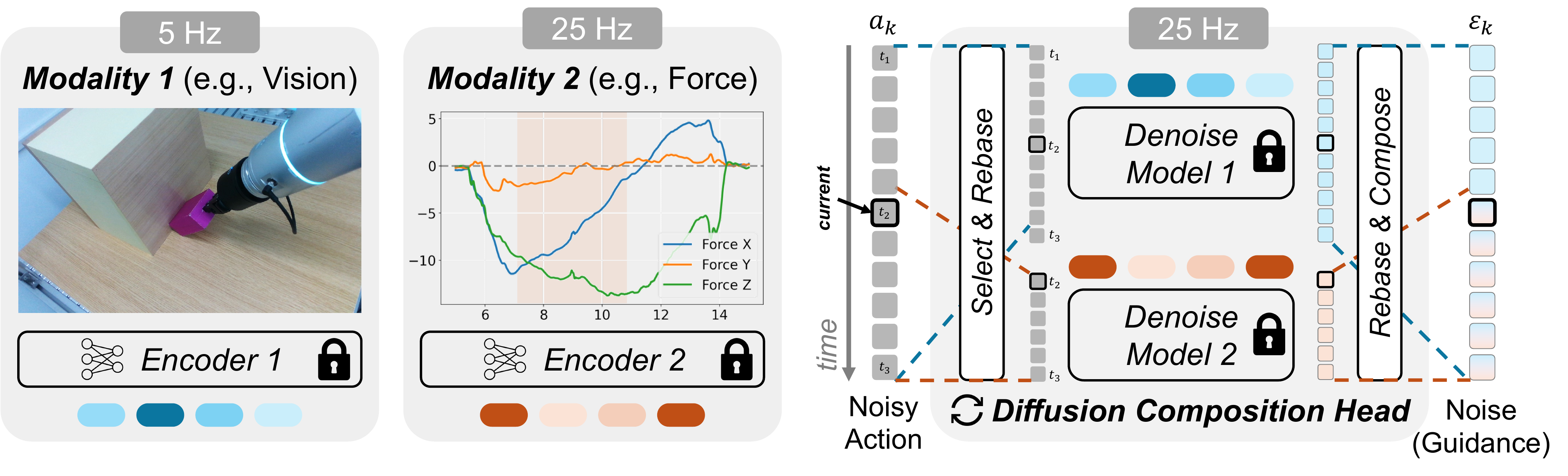}
    \caption{\textbf{Overview of the \mate{} Framework.}
    \mate{} asynchronously composes component diffusion policies operating at heterogeneous inference rates.
    \textbf{\textit{(Left)}} We illustrate vision-force composition as an example, where a lower-rate vision-conditioned policy provides long-horizon scene guidance and a higher-rate force-conditioned policy provides local contact guidance.
    \textbf{\textit{(Right)}} In the diffusion composition head, the overlapping noisy action segment is rebased into the current force frame for force denoising. The corresponding vision guidance is rebased to the same frame, fused with force guidance, and mapped back to update the full-horizon noise prediction.}
    \label{fig:model}
    \vspace{-1.0em}
\end{figure*}

\subsection{Policy Inference Frequency and Multi-Rate Execution}
Policy inference frequency determines how often new observations can influence action predictions, and is distinct from sensor sampling and low-level control rates.
Prior work improves policy responsiveness through consistency distillation~\cite{prasad2024consistency, lu2024manicm} or by reusing previous predictions to reduce denoising computation~\cite{hoeg2025streaming, duan2025realtime}.
Other approaches address execution latency through observation-action latency matching~\cite{umi} or asynchronous action chunk generation with continuity constraints~\cite{rtc}.
These methods improve inference efficiency and execution timing for individual policies.

Multi-rate architectures provide another way to improve responsiveness by operating different components at different temporal scales.
This is particularly useful in contact-rich manipulation, where rapidly changing interaction forces can require frequent feedback-driven corrections~\cite{rdp, foar, forcepolicy, acp, forcemimic}.
For example, Reactive Diffusion Policy (RDP) combines a slow latent diffusion policy with a faster tactile-conditioned decoder~\cite{rdp}, while Force Policy uses a slower global vision policy together with a higher-rate local force policy~\cite{forcepolicy}.
Related slow-fast and hierarchical architectures likewise employ lightweight reactive components between slower high-level updates~\cite{hirt, fastinslow, trex}.
These works demonstrate the benefit of heterogeneous inference rates, but their components are typically designed and trained as coupled architectures rather than independently reusable policies.
\mate{} combines this rate-preserving capability with the modular policy composition setting discussed above, allowing independently trained policies to contribute asynchronously at their native inference rates without being bottlenecked by the slowest component.

\section{Method}

We propose \mate{}, a general framework for asynchronous multi-rate diffusion policy composition.
The framework composes independently trained component policies that may use different sensing modalities, observation representations, and inference rates, without synchronizing their execution to the slowest component.
Each policy is queried according to its own inference schedule, allowing newly available guidance from a faster policy to update the composed action while reusing guidance from slower policies.
Fig.~\ref{fig:model} illustrates a vision-force instantiation: a lower-rate vision policy provides global, long-horizon guidance, while a higher-rate force/torque policy supplies local contact feedback.

At inference time, \mate{} maintains the latest denoising guidance from each component policy and composes their temporally overlapping predictions during diffusion sampling.
Unlike synchronous composition (\S\ref{sec:method-preliminary}), \mate{} aligns predictions from different query times by rebasing diffusion variables into a common action reference frame (\S\ref{sec:method-rebase}), then composes the aligned guidance over temporally overlapping horizons (\S\ref{sec:method-fusion}).

\subsection{Preliminary: Synchronous Diffusion Policy Composition}\label{sec:method-preliminary}
We first review the standard setting of diffusion policy composition~\cite{gpc,poco}.
Let $a_0 \in \mathbb{R}^{H \times d}$ denote a clean action trajectory over horizon $H$ and dimension $d$. A diffusion policy defines a noisy action as
\begin{equation}
    a_k
    =
    \sqrt{\bar{\alpha}_k} a_0
    +
    \sqrt{1-\bar{\alpha}_k}\epsilon,
    \qquad
    \epsilon \sim \mathcal{N}(0,I),
    \label{eq:forward_diffusion}
\end{equation}
% where $k$ is the diffusion timestep and $\bar{\alpha}_k$ is the cumulative noise schedule. 
% Given a noisy action $a_k$, a modality-conditioned policy predicts the noise term under its own observation.
% For two modalities in this work, we have $\epsilon^{V}_k=\pi^{V}\left(a_k,k;c^{V}\right)$ for vision and 
% $\epsilon^{F}_k=\pi^{F}\left(a_k,k;c^{F}\right)$ for force,
% where $c^{V}$ and $c^{F}$ denote the corresponding modality conditions after observation encoding.

% In the synchronous setting, both predictions are evaluated at the same physical time and in the same action reference frame.
% Therefore, they can be directly composed as
% \begin{equation}
%     \epsilon_k =
%     \frac{w^V}{w^V+w^F} \epsilon^V_k
%     +
%     \frac{w^F}{w^V+w^F} \epsilon^F_k,
%     \label{eq:basic_noise_composition}
% \end{equation}
% where $w^V,w^F\ge0$ are the composition weights for each modality. The composed noise prediction is then used in a standard DDIM~\cite{ddim} update $a_{k-1}=\operatorname{DDIMStep}\left(a_k, \epsilon_k, k\right)$. This synchronous composition serves as the basis for our asynchronous extension.

where $k$ is the diffusion timestep and $\bar{\alpha}_k$ is the cumulative noise schedule. 
Consider a collection of $M$ component diffusion policies,${\pi^{(m)}}_{m=1}^{M}$, that share a compatible action space and diffusion parameterization.Each policy may use its own observation representation and conditioning pipeline.
Given the shared noisy trajectory $a_k$, policy $m$ predicts
\begin{equation}
\epsilon_k^{(m)}
=\pi^{(m)}\left(a_k,k;c^{(m)}\right)
\qquad m \in {1,\ldots,M},\label{eq:component-guidance}
\end{equation}
where $c^{(m)}$ denotes the policy-specific condition after observation encoding.
Component policies may use different sensing modalities, distinct representations of the same modality, or complementary subsets of the available observations.

In the synchronous setting, all component policies are queried at the same physical time, and their predictions are defined over the same action horizon and in the same action reference frame.
Their noise predictions can therefore be directly composed as
\begin{equation}
\epsilon_k=\sum_{m=1}^{M}\frac{w^{(m)}}{\sum_{j=1}^{M}w^{(j)}}\epsilon_k^{(m)},\label{eq:synchronous-composition}
\end{equation}

where $w^{(m)} \geq 0$ is the composition weight of the $m$-th policy and $\sum_{m=1}^{M}w^{(m)} > 0$.
The composed noise prediction is then used in a standard DDIM~\cite{ddim} update $a_{k-1}=\operatorname{DDIMStep}\left(a_k, \epsilon_k, k\right)$. This synchronous composition serves as the basis for our asynchronous extension.

\subsection{Reference-Frame Rebase for Diffusion Variables}\label{sec:method-rebase}

% In asynchronous composition, all modality guidance must be expressed in a common action reference frame.
% We use relative action chunks~\cite{cage,diffusionpolicy,umi,act}, where each action is defined with respect to the robot state at the policy query time.
% This representation is important for local feedback policies, since force-conditioned actions should describe corrections relative to the current contact state rather than absolute target poses~\cite{forcepolicy}.
% However, relative actions are tied to their starting state: the same relative action can lead to different physical motions when applied from different robot poses.
% Therefore, when a low-frequency policy is queried at time $t_1$ and reused at a later time $t_2$, its predicted diffusion variables are still expressed in the action frame at $t_1$.
% Directly fusing them with guidance produced at $t_2$ would mix predictions from different reference frames.
% We address this by rebasing delayed diffusion variables into the current action frame before fusion, making clean actions, noisy samples, and predicted noises comparable across asynchronous diffusion policies.

In asynchronous composition, the component policies may be queried at different physical times. Consequently, their predicted action trajectories and denoising guidance may be expressed in different action reference frames. Before composition, all available diffusion variables must therefore be transformed into a common reference frame.

We use relative action chunks~\cite{cage,diffusionpolicy,umi,act}, where each predicted action is defined with respect to the robot state at the corresponding policy query time. Relative actions provide a convenient representation for closed-loop control, but they are inherently tied to their initial reference state: the same relative action can induce different physical motions when executed from different robot poses. Therefore, directly composing predictions generated at different query times would mix diffusion variables expressed in different action frames.

Consider an arbitrary pair of policy query events at times $t_1$ and $t_2$, with $t_1 \leq t_2$. A cached prediction generated at $t_1$ is expressed in the action frame associated with the robot pose at $t_1$, whereas newly available guidance at $t_2$ is expressed in the frame associated with the robot pose at $t_2$. We derive a pairwise rebase operation that transforms diffusion variables from the source frame at $t_1$ to the target frame at $t_2$. This operation is independent of the observation modality or policy architecture and can be applied separately to every cached policy prediction before composition.

Let $T_1$ and $T_2$ denote the robot poses at $t_1$ and $t_2$, respectively. Then, the transformation from the old action frame at $t_1$ to the current action frame at $t_2$ is
$T_{2 \leftarrow 1}=T_2^{-1}T_1$. Let $T_{2\leftarrow 1}=\left(R_{2 \leftarrow 1}, p_{2 \leftarrow 1}\right)$, where $R_{2 \leftarrow 1}$ denotes the rotation matrix and $p_{2 \leftarrow 1}$ is the translational offset.

\begin{figure}[t]
    \centering
    \includegraphics[width=\columnwidth]{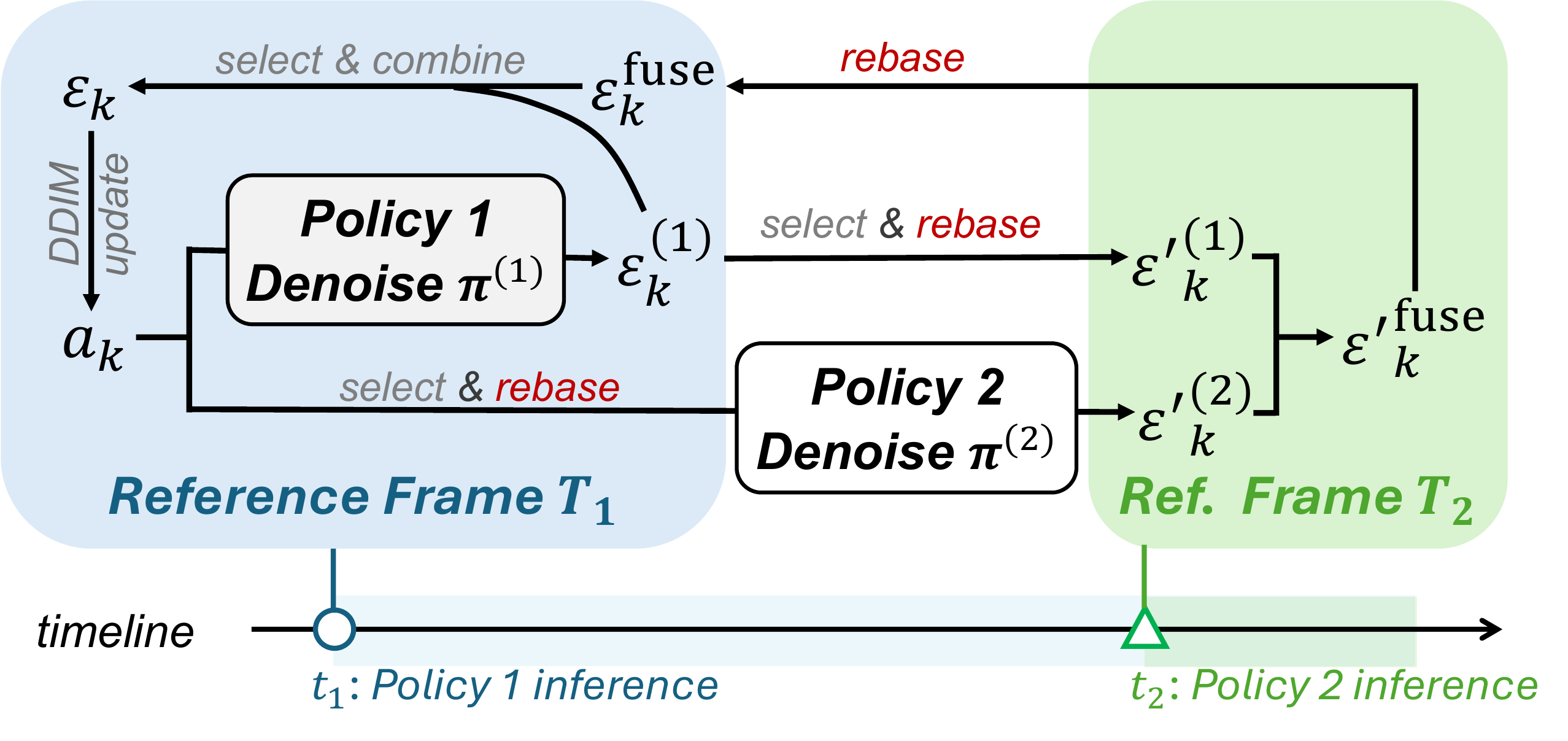}
    \caption{\textbf{Pairwise Asynchronous Policy Composition with Reference-Frame Rebasing.}
    We illustrate the composition of two component policies queried at different times.
    Policy~1 provides full-horizon guidance in frame $T_1$, while Policy~2 is queried later and produces guidance in frame $T_2$.
    \mate{} selects the temporally overlapping noisy action segment and rebases it from $T_1$ to $T_2$ for Policy~2 denoising.
    The corresponding Policy~1 guidance is also selected and rebased to $T_2$, where the aligned guidance from both policies is fused.
    The fused guidance is then mapped back to $T_1$ and inserted into the full-horizon diffusion prediction.}
    \label{fig:fusion}
    \vspace{-0.4cm}
\end{figure}

% \paragraph{Action Representation and Transformation.} 
\textbf{Action Representation and Transformation.} Since composition is performed in the noise space, the rebase operation must be compatible with clean actions, noisy samples, and predicted noises.
We thus need an action representation whose frame transformation has an orthogonal linear part and a separate affine offset.
The linear part can be applied to noises without changing their Gaussian covariance, while the offset must be handled separately for clean actions and noisy samples.

We first consider the rotation component.
Let the rotation component of a relative action with respect to frame $T_1$ be
$R=[r_1,r_2,r_3]\in SO(3)$, where $r_i\in\mathbb{R}^3$ is the $i$-th column.
Under the transformation $T_{2\leftarrow 1}$, the same relative rotation is expressed in frame $T_2$ as $R'=R_{2\leftarrow 1}R$, so we have $r'_i=R_{2\leftarrow 1}r_i$.
This column-wise transformation motivates the column-based 6D representation~\cite{rotation_6d}, $R_\mathrm{6d}=[r_1^\top,r_2^\top]^\top$, whose rebase is linear:
$
    R'_{\mathrm{6d}}
    =
    \operatorname{blockdiag}
    \left(
        R_{2\leftarrow 1},
        R_{2\leftarrow 1}
    \right)\cdot 
    R_{\mathrm{6d}}
$.
Other action dimensions can also be handled in the same representation space.
For the relative translation $x\in\mathbb{R}^{3}$, the clean-action rebase is affine: $x'=R_{2\leftarrow 1}x+p_{2\leftarrow 1}$.
The gripper command $g\in\mathbb{R}$ is frame-invariant, so $g'=g$.
Thus, each action step is represented as
$a=[x^\top,R_{\mathrm{6d}}^\top,g]^\top\in\mathbb{R}^{10}$, and the full action-space transformation is
\begin{equation}
    a' = A_{2\leftarrow 1} a + b_{2\leftarrow 1},
\end{equation}
where
\begin{equation}
    \left\{\begin{aligned}
    A_{2\leftarrow 1}
    &=
    \operatorname{blockdiag}
    \left(
        R_{2\leftarrow 1},
        R_{2\leftarrow 1},
        R_{2\leftarrow 1},
        1
    \right)\in\mathbb{R}^{10\times 10},\\b_{2\leftarrow 1}&=\left[
        p_{2\leftarrow 1},
        \mathbf{0}_{3}^\top,
        \mathbf{0}_3^\top,
        0
    \right]^\top \in \mathbb{R}^{10}. \end{aligned}\right.
    \label{eq:full_action_rebase_matrix}
\end{equation}
Since $R_{2\leftarrow 1}\in SO(3)$, the linear part $A_{2\leftarrow 1}$ is orthogonal.
Next, we show that this affine rebase leads to different rules for different diffusion variables: clean actions use the full affine transform, noisy samples include a scaled offset, and noises are transformed only by the orthogonal linear part $A_{2\leftarrow 1}$.

\textbf{Diffusion Variables Transformation.} We now derive how the affine rebase applies to different diffusion variables. For the clean actions, they naturally follow the full affine transform in Eq.~\eqref{eq:full_action_rebase_matrix}. 
Given a noisy sample $a_k$ and a predicted noise $\epsilon_k$ in frame $T_1$, the clean action estimate is
\begin{equation}
    \hat{a}_{0}
    =
    \frac{1}{\sqrt{\bar{\alpha}_k}}a_{k}
    -
    \frac{\sqrt{1-\bar{\alpha}_k}}{\sqrt{\bar{\alpha}_k}}
    \epsilon_{k}.
    \label{eq:x0_prediction_t1}
\end{equation}
Applying the clean action rebase in Eq.~\eqref{eq:full_action_rebase_matrix} gives
\begin{equation}
\begin{aligned}
    \hat{a}'_{0}
    &= A_{2\leftarrow 1}\hat{a}_{0} + b_{2\leftarrow 1} \\
    &= \frac{1}{\sqrt{\bar{\alpha}_k}}
    \left(
        A_{2\leftarrow 1}a_{k}
        +
        \sqrt{\bar{\alpha}_k}b_{2\leftarrow 1}
    \right)
    -
    \frac{\sqrt{1-\bar{\alpha}_k}}{\sqrt{\bar{\alpha}_k}}
    A_{2\leftarrow 1}\epsilon_{k}.
\end{aligned}
\label{eq:rebase_x0_derivation}
\end{equation}
Comparing Eq.~\eqref{eq:rebase_x0_derivation} with the standard estimator in frame $T_2$, which has the same form as Eq.~\eqref{eq:x0_prediction_t1}, we can obtain the rebase rules:
\begin{equation}
\left\{
\begin{aligned}
    a'_{k}
    &=
    A_{2\leftarrow 1}a_{k}
    +
    \sqrt{\bar{\alpha}_k}b_{2\leftarrow 1},
    \\
    \epsilon'_{k}
    &=
    A_{2\leftarrow 1}\epsilon_{k}.
\end{aligned}
\right.
\label{eq:noise_transform}
\end{equation}
Thus, the affine offset appears in the noisy sample with a factor $\sqrt{\bar{\alpha}_k}$, but it is not added to the diffusion noise.
The noise is transformed only by the orthogonal linear part $A_{2\leftarrow 1}$, preserving the standard Gaussian diffusion distribution.

The derivation above defines a pairwise rebase operator between two action frames and does not assume a particular policy or observation modality. More generally, suppose that composition is performed at time $t_c$ in the current action frame $T_c$, and let $\tau_m \leq t_c$ denote the query time of the most recent available prediction from component policy $\pi^{(m)}$. Its prediction is associated with the action frame $T_m=T(\tau_m)$. We independently compute $T_{c\leftarrow m} = T_c^{-1}T_m$
and apply the rebase rules in Eq.~\eqref{eq:noise_transform} to transform the corresponding diffusion variables into $T_c$. After this operation is applied to every available component prediction, all guidance terms are expressed in the same action reference frame and can be composed consistently.

\subsection{Asynchronous Multi-Rate Policy Composition}
\label{sec:method-fusion}

\mate{} supports the asynchronous composition of $M$ component diffusion policies
$\{\pi^{(m)}\}_{m=1}^{M}$ operating at heterogeneous inference rates.
At a composition time $t_c$, the latest available prediction from each policy may have been generated at a different query time, over a different temporal horizon, and in a different action reference frame.
We first derive asynchronous fusion for an arbitrary pair of component policies and then extend the formulation to multi-policy composition.

\textbf{Pairwise asynchronous fusion.}
Consider two arbitrary component policies $\pi^{(1)}$ and $\pi^{(2)}$, queried at times $t_1$ and $t_2$, respectively, where $t_1<t_2$.
Their predictions are expressed in the corresponding action frames $T_1$ and $T_2$.
% Without loss of generality, suppose that $\pi^{(1)}$ provides a longer-horizon prediction, while the prediction horizon of $\pi^{(2)}$ overlaps with a segment of the first prediction.
Fig.~\ref{fig:fusion} illustrates the asynchronous composition of these two arbitrary component policies.
Policy $\pi^{(1)}$ provides full-horizon guidance in frame $T_1$, while Policy $\pi^{(2)}$ is queried later and produces guidance over an overlapping horizon in frame $T_2$.
The following derivation applies to any pair of component policies with compatible action spaces and overlapping prediction horizons, regardless of their observation modalities, representations, or inference rates.

% We now instantiate the general formulation with two component policies:
% a lower-rate vision-conditioned policy $\pi^{V}$ and a higher-rate force-conditioned policy $\pi^{F}$.
% This example is illustrated in Fig.~\ref{fig:fusion}.
% Suppose the vision policy is queried at time $t_1$, and the force policy is queried later at time $t_2>t_1$.
% We maintain the full-horizon noisy action $a_k\in\mathbb{R}^{H\times d}$ in the vision action frame $T_1$, and assume that the force horizon is contained within the vision horizon.

% We now apply the rebase rule to asynchronous policy composition, as illustrated in Figure~\ref{fig:fusion}.
% In our instantiation, the vision policy is queried at a lower frequency and provides full-horizon guidance, while the force policy is queried at a higher frequency and provides local contact-aware guidance.
% Suppose the vision policy is queried at time $t_1$, and the force policy is queried later at time $t_2>t_1$.
% We maintain the full-horizon noisy action $a_k\in\mathbb{R}^{H\times d}$ in the vision action frame $T_1$, and assume that the force horizon is contained within the vision horizon.

Let $S(\cdot)$ denote the temporal selection operator that extracts the segment of Policy~1's prediction overlapping with the horizon of Policy~2.
Before querying Policy~2, we rebase this selected noisy segment into Policy~2's action frame $T_2$ using Eq.~\eqref{eq:noise_transform}.
Policy~2 then predicts
\begin{equation}
    \epsilon'^{(2)}_k
    =
    \pi^{(2)}
    \left(
        A_{2\leftarrow 1}S(a_k)
        +
        \sqrt{\bar{\alpha}_k}b_{2\leftarrow 1},
        k;
        c^{(2)}
    \right),
    \label{eq:force_guidance_prediction}
\end{equation}
where $c^{(2)}$ is Policy~2's condition at time $t_2$. The rebase parameters are applied to every action step.

% Let $S(\cdot)$ denote the temporal selection operator that extracts the segment of the vision horizon corresponding to the force horizon.
% Before querying the force policy, we rebase this selected noisy segment into the force action frame $T_2$ using Eq.~\eqref{eq:noise_transform}.
% The force policy then predicts
% \begin{equation}
%     \epsilon'^F_k
%     =
%     \pi^F
%     \left(
%         A_{2\leftarrow 1}S(a_k)
%         +
%         \sqrt{\bar{\alpha}_k}b_{2\leftarrow 1},
%         k;
%         c^F
%     \right),
%     \label{eq:force_guidance_prediction}
% \end{equation}
% where $c^F$ is the force condition at time $t_2$. The rebase parameters are applied to every action step.

Policy~1 predicts full-horizon guidance $\epsilon^{(1)}_k=\pi^{(1)}(a_k,k;c^{(1)})$ in frame $T_1$, where $c^{(1)}$ is Policy~1's condition at time $t_1$.
To fuse it with Policy~2's guidance, we select the same temporal segment and rebase Policy~1's guidance into frame $T_2$:
\begin{equation}
    \epsilon'^{(1)}_k
    =
    A_{2\leftarrow 1}S(\epsilon^{(1)}_k).
    \label{eq:vision_guidance_to_force_frame}
\end{equation}

% The vision policy predicts full-horizon guidance $\epsilon^V_k=\pi^V(a_k,k;c^V)$ in frame $T_1$, where $c^V$ is the vision condition at time $t_1$.
% To fuse it with the force guidance, we select the same temporal segment and rebase the vision guidance into frame $T_2$:
% \begin{equation}
%     \epsilon'^V_k
%     =
%     A_{2\leftarrow 1}S(\epsilon^V_k).
%     \label{eq:vision_guidance_to_force_frame}
% \end{equation}

\textbf{Extension to multiple component policies.}
The transformation above is pairwise and does not depend on the observation modality or policy architecture.
It can therefore be independently applied to every component policy whose prediction participates in the current composition.
At composition time $t_c$, we choose a common target frame $T_c$, typically the action frame associated with the most recently queried policy.
Let $\tau^{(m)}\leq t_c$ denote the query time of the latest available prediction from policy $\pi^{(m)}$, and let
$T_m=T(\tau^{(m)})$ denote its associated action frame.
For each policy, we select the portion of its guidance that overlaps with the composed horizon and independently rebase it into $T_c$:
\begin{equation}
\epsilon'^{(m)}_k
=
\mathbf{A}_{c\leftarrow m}
S(\epsilon^{(m)}_k),
\qquad m\in\mathcal{M}_c,
\label{eq:multi-policy-rebase}
\end{equation}
where $\mathcal{M}_c$ is the set of component policies available for composition at $t_c$.
After this transformation, all selected guidance terms share the same temporal support and action reference frame.

The aligned guidance terms can then be composed through a normalized weighted combination:
\begin{equation}
\epsilon^{\mathrm{fuse}}_{k,i}
=
{
\displaystyle
\sum_{m\in\mathcal{M}_{c,i}}
w_i^{(m)}
\epsilon'^{(m)}_{k,i}
}/{
\displaystyle
\sum_{m\in\mathcal{M}_{c,i}}
w_i^{(m)}
},
\label{eq:multi-policy-fusion}
\end{equation}
where $\mathcal{M}_{c,i}\subseteq\mathcal{M}_c$ contains the policies whose predictions cover horizon step $i$, and $w_i^{(m)}\geq0$ denotes the contribution of policy $m$ at that step.
The weights can reflect prediction freshness, horizon position, or component reliability.

Finally, we map the fused guidance from the common frame $T_c$ back to the original full-horizon action frame, replace the corresponding segment of the full-horizon guidance while leaving the rest unchanged, and use the updated guidance for the DDIM update.

\section{Experiments}
\label{sec:result}

\begin{figure*}[t]
    \centering
    \includegraphics[width=\textwidth]{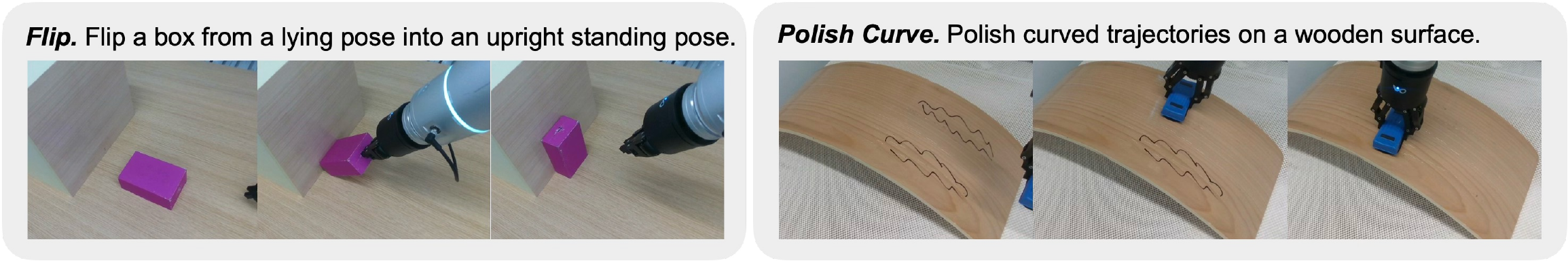}
    \caption{\textbf{Contact-Rich Manipulation Tasks.}
    We evaluate \mate{} on two contact-rich tasks that require both visual perception and fine-grained force/torque feedback. Both tasks require effective coordination and adaptation across visual and force/torque modalities.}
    \label{fig:task}
    \vspace{-0.1cm}
\end{figure*}

\begin{figure*}[t]
\centering

% =========================================================
% Left: Main evaluation table
% =========================================================
\begin{minipage}[t]{0.65\textwidth}
\vspace{0pt}
\centering
\footnotesize
\setlength{\tabcolsep}{3pt}

\resizebox{\linewidth}{!}{%
\begin{tabular}{lcc|cccc}
\toprule
\multirow{2}{*}{\textbf{Policy}}
& \multirow{2}{*}{\textbf{Modular}}
& \multirow{2}{*}{\makecell{\textbf{Inference}\\\textbf{Freq.}}}
& \multicolumn{2}{c}{\textbf{\textit{Flip}}}
& \multicolumn{2}{c}{\textbf{\textit{Polish Curve}}} \\
\cmidrule(lr){4-5}
\cmidrule(lr){6-7}
& & & Push & Flip & Contact & Score \\
\midrule

RISE~\cite{rise} (vision policy)
& - & 5 Hz
& 95.0\% & 30.0\% & 90.0\% & 35.00\% \\

Force-only
& - & \cellcolor{highfreq}25 Hz
& 5.0\% & 5.0\% & 25.0\% & 6.25\% \\

\midrule

Training Noise Fusion
& \xmark & 5 Hz
& 30.0\% & 17.5\% & 70.0\% & 20.00\% \\

Synchronous Composition
& \checkmark & 5 Hz
& \textbf{100.0\%} & 25.0\%
& \textbf{100.0\%} & 25.00\% \\

Policy Consensus~\cite{multimodal_consensus}
& \checkmark$^\dagger$ & 5 Hz
& 95.0\% & 40.0\%
& \textbf{100.0\%} & 42.50\% \\

Absolute Action
& \checkmark & \cellcolor{highfreq}25 Hz
& 60.0\% & 12.5\% & 75.0\% & 21.25\% \\

Relative Action w/o Rebasing
& \checkmark & \cellcolor{highfreq}25 Hz
& 90.0\% & 47.5\%
& \textbf{100.0\%} & 37.50\% \\

\midrule

TA-VLA~\cite{tavla}
& \xmark & 5 Hz
& 80.0\% & 30.0\%
& \textbf{100.0\%} & 47.50\% \\

RDP~\cite{rdp}
& \xmark & \cellcolor{highfreq}24 Hz
& \textbf{100.0\%} & 35.0\%
& \textbf{100.0\%} & 45.00\% \\

\midrule

\mate{} (\textit{ours})
& \checkmark & \cellcolor{highfreq}25 Hz
& 95.0\% & \textbf{60.0\%}
& \textbf{100.0\%} & \textbf{55.00\%} \\

\bottomrule
\end{tabular}%
}

\vspace{0.1cm}
\captionof{table}{\textbf{Evaluation Results.}
\mate{} achieves the best final-task performance, demonstrating the effectiveness of modular, rate-preserving policy composition.
Gray cells indicate high-frequency inference.
The $\dagger$ denotes that Policy Consensus~\cite{multimodal_consensus} requires training an additional router and fine-tuning the component policies.}
\label{tab:flip_polish_curve_results}
\end{minipage}
\hfill
% =========================================================
% Right: Frequency figure
% =========================================================
\begin{minipage}[t]{0.33\textwidth}
\vspace{0pt}
\centering

\includegraphics[width=\linewidth]
{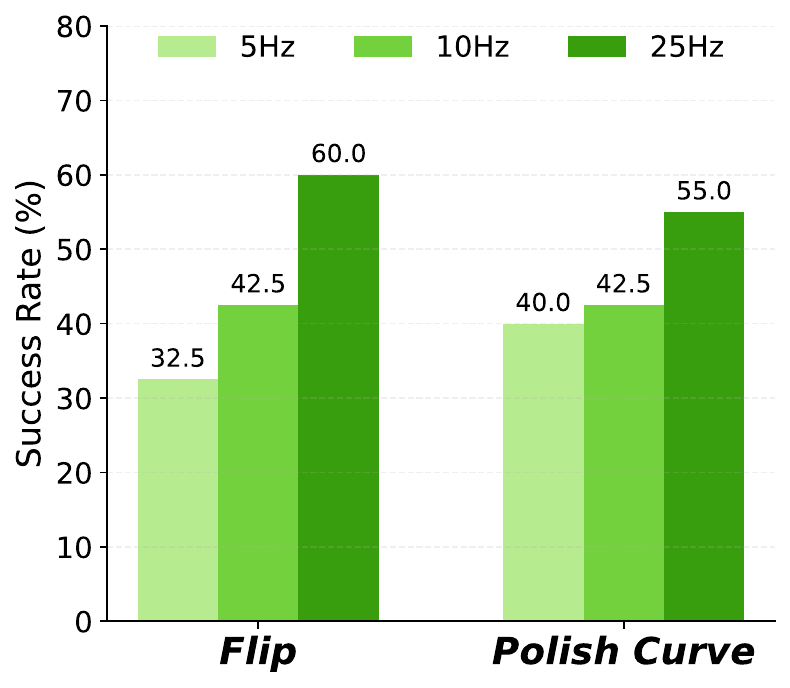}

\vspace{0.1cm}
\captionof{figure}{\textbf{Effect of Inference Rate.}
Final-task performance on \textit{Flip} and \textit{Polish Curve} at different inference rates.
Higher rates enable more timely guidance and improve the composed policy on both tasks.}
\label{fig:freq}
\end{minipage}

\vspace{-0.4cm}
\end{figure*}

We answer the following research questions through experiments:
\textbf{(Q1)} Does asynchronous multi-rate composition improve performance over existing composition strategies and policy architectures?
\textbf{(Q2)} What roles do relative actions and reference-frame rebasing play in asynchronous composition?
\textbf{(Q3)} Can \mate{} effectively compose heterogeneous policies across representations, viewpoints, and inference rates?
\textbf{(Q4)} How does preserving policy inference rate affect composed-policy performance?

% We answer the following research questions through experiments:
% \textbf{(Q1)} Does \mate{} outperform synchronous composition and existing vision-only and multimodal policies, including a matched-frequency baseline?
% \textbf{(Q2)} Are relative action representations and reference-frame rebasing necessary for asynchronous composition?
% \textbf{(Q3)} Can \mate{} generalize beyond vision-force fusion to visual policies with heterogeneous representations and viewpoints?
% \textbf{(Q4)} How does inference rate affect \mate{} performance?

\subsection{Setup}
\begin{figure*}[t]
    \centering
    \includegraphics[width=\textwidth]{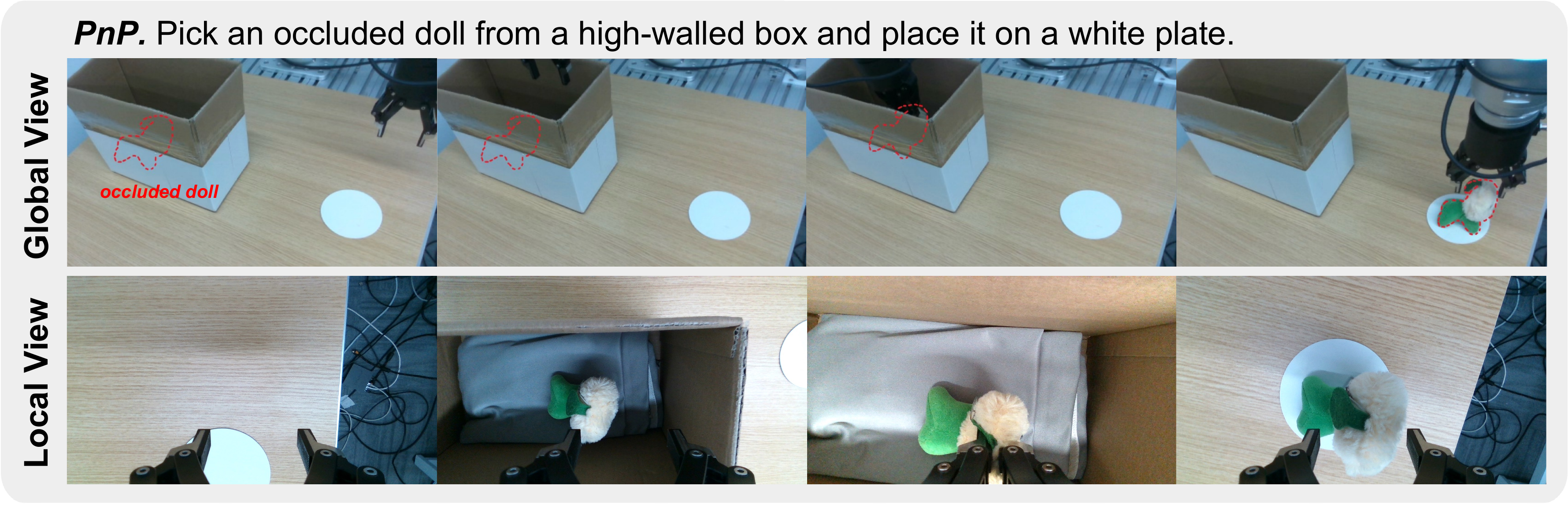}
    \caption{\textbf{Occluded Pick and Place (\textit{PnP}) Task.}
    The robot reaches into a high-walled box, grasps a doll, and places it on a designated white disk. The box walls frequently occlude the doll from the global camera, requiring complementary close-range observations from the wrist camera.}
    \label{fig:task-pnp}
    \vspace{-0.4cm}
\end{figure*}

\textbf{Tasks.}
We evaluate \mate{} on three real-world manipulation tasks. The \textit{Flip} and \textit{Polish Curve} tasks, shown in Fig.~\ref{fig:task}, require both global visual perception and fine-grained force feedback. The Occluded Pick-and-Place (\textit{PnP}) task, shown in Fig.~\ref{fig:task-pnp}, requires the robot to retrieve a occluded doll from a high-walled box and place it on a designated white disk.
For \textit{Flip} and \textit{Polish Curve}, we instantiate \mate{} by composing a RISE-based vision policy with a lightweight force policy, operating at inference rates of 5~Hz and 25~Hz, respectively.
For \textit{PnP}, we compose a global-camera 3D vision policy based on RISE with a wrist-camera 2D vision policy, operating at 2~Hz and 5~Hz, respectively.
These component inference rates are used unless otherwise stated.

\textbf{Data Collection.}
We perform real-world experiments on a Flexiv Rizon4 robot equipped with an FT03S six-axis force/torque sensor.
For each task, we collect 50 demonstrations using an arm-to-arm teleoperation system with accurate force feedback~\cite{tdk}.
This setup allows the operator to provide contact-rich manipulation trajectories while preserving fine-grained interaction information between the robot and the environment.
During data collection, we record synchronized multimodal observations, including color images, depth images, robot proprioceptive states, and force/torque measurements.
The color and depth streams are recorded at 20 Hz, while the low-dimensional data is recorded at 1000 Hz.

\textbf{Baselines.} 
% To isolate the contribution of the composition design, we include five critical controls: Training-Time Noise Fusion, Synchronous Composition, Policy Consensus~\cite{multimodal_consensus}, Absolute Action, and Relative Action w/o Rebasing.
To isolate the contribution of the composition design, we include five critical controls.
Training-Time Noise Fusion combines diffusion noises during training, whereas Synchronous Composition combines modality-specific policy branches at synchronized inference steps.
Building on Synchronous Composition, Policy Consensus~\cite{multimodal_consensus} employs a learned router to determine the contribution of each modality.
Absolute Action asynchronously composes policies that directly predict actions in a shared absolute frame.
Relative Action w/o Rebasing instead represents actions relative to each policy's latest observation but directly fuses these predictions without transforming them to a common temporal reference frame.
We further compare against the representative multimodal policies TA-VLA~\cite{tavla} and RDP~\cite{rdp}.

\textbf{Metrics.}
For \textit{Flip}, ``Push'' measures whether the box reaches the target, and ``Flip'' measures successful flipping, with partial flips counted as half successes.
For \textit{Polish Curve}, ``Contact'' measures surface contact, and ``Score'' is $1.0/0.75/0.25/0$ for complete erasure, over-$50\%$ erasure, partial erasure below $50\%$, and failure, respectively.
For \textit{PnP}, ``Reach'' measures whether the robot's end effector enters the high-walled box, and ``Pick'' measures whether the gripper grasps the toy. ``Place'' measures whether the robot successfully deposits the toy on the designated white disk.

\textbf{Protocols.}
All policies are deployed on a local workstation equipped with a NVIDIA RTX 5090 GPU, with the exception of TA-VLA~\cite{tavla}, whose memory footprint exceeds the workstation's GPU capacity. 
TA-VLA~\cite{tavla} is therefore served on a remote node with 4 NVIDIA A100 GPUs. 
To ensure a fair and reproducible comparison, we adopt an identical evaluation protocol across all policies. We pre-sample a fixed set of initial object placements that are uniformly distributed over the workspace, and reuse the same set of placements for every policy. 
This isolates the effect of the policy itself from variation in the test layout, so that all methods are evaluated under identical initial conditions. 
Unless otherwise stated, the main results use the seen object and randomize only its placement. Each policy is evaluated over 20 consecutive trials per task.

\subsection{Vision-Force Policy Composition}

% \textbf{Asynchronous latency-aware guidance fusion improves multimodal composition performance (Q1).}
% \textbf{Asynchronous multi-rate composition improves over synchronous composition baselines (Q1).}
% \textbf{Asynchronous multi-rate composition outperforms synchronous and frequency-agnostic composition methods (Q1).}
\textbf{Asynchronous multi-rate composition improves over existing composition strategies (Q1).}
Table~\ref{tab:flip_polish_curve_results} shows that \mate{} consistently outperforms other fusion methods. 
For training-time noise fusion, high-frequency force inputs are forced into temporal alignment with slower vision observations, limiting their capability.
Compared with synchronous composition, this gap suggests that bottlenecking all modalities at slow inference frequency limits the utilization of modality-specific capabilities, especially for lightweight, high-frequency modalities such as force feedback.
Policy Consensus further adds a router on top of synchronous composition, but still underperforms \mate{}. We found that the learned router weights may overfit to one dominant modality during training, leading to imbalanced modality usage.
Instead, \mate{} asynchronously composes modality-specific policies at inference time, allowing each modality branch to operate at its native frequency rather than being bottlenecked by the slowest modality.

% \textbf{\mate{} achieves stronger overall performance than existing vision-only and multimodal baselines (Q2).}
% \textbf{\mate{} achieves stronger overall performance than existing vision-only and multimodal policies (Q1).}
\textbf{\mate{} achieves stronger overall performance than existing policy architectures (Q1).}
Standalone and jointly trained policy architectures exhibit different limitations on these contact-rich tasks.
The vision-only RISE backbone lacks contact-aware feedback and struggles during contact-sensitive stages.
RDP adopts a specially designed jointly trained multimodal architecture, but does not explicitly align and compose modality-specific guidance at inference time.
TA-VLA is limited by its low inference speed, which restricts reactive correction using force feedback.
% TA-VLA incorporates tactile feedback but operates at a 5\,Hz action-update rate under its action-chunking execution scheme, limiting the frequency of closed-loop correction.
By contrast, \mate{} combines long-horizon visual guidance with high-frequency force feedback, enabling more responsive contact adaptation and stronger overall task performance.

% \textbf{Relative actions and reference-frame rebasing are both necessary for asynchronous composition.}
% \textbf{Relative actions and reference-frame rebasing are necessary for asynchronous composition (Q2).}
\textbf{Relative actions and reference-frame rebasing enable consistent asynchronous composition (Q2).}
Predicting absolute actions substantially degrades performance because the force-only policy receives no proprioceptive state: force observations describe local contact but do not determine the robot's global end-effector pose, making absolute action prediction ambiguous.
We therefore predict actions relative to the end-effector frame for all policies.
However, asynchronous policies produce these relative actions at different robot poses, so directly averaging them without rebasing composes geometrically inconsistent guidance.
Although this operation is mathematically inconsistent, its empirical degradation is moderate at the tested 5 Hz vision policy frequency because the robot typically undergoes limited motion during the short interval, resulting in a relatively small reference-frame mismatch (Table~\ref{tab:flip_polish_curve_results}).
Rebasing transforms the asynchronous predictions into a common current reference frame before fusion, enabling consistent multi-rate composition.

\subsection{3D-2D Vision Policy Composition}

\begin{table}[t]
\centering
\footnotesize
\setlength{\tabcolsep}{4pt}

\begin{tabular}{lc|ccc}
\toprule
\multirow{2}{*}{\textbf{Policy}}
& \multirow{2}{*}{\makecell{\textbf{Infer}\\\textbf{Freq.}}}
& \multicolumn{3}{c}{\textbf{\textit{PnP}}} \\
\cmidrule(lr){3-5}
& & Reach & Pick & Place \\
\midrule

RISE~\cite{rise}
& 2 Hz
& 60.0\% & 15.0\% & 10.0\% \\

\midrule

Training Noise Fusion
& 2 Hz
& 95.0\% & 85.0\% & 80.0\% \\

Sync. Composition
& 2 Hz
& 90.0\% & 80.0\% & 75.0\% \\

Policy Consensus~\cite{multimodal_consensus}
& 2 Hz
& \textbf{100.0\%} & 85.0\% & 80.0\% \\

\midrule

\mate{} (\textit{ours})
& \cellcolor{highfreq}5 Hz
& \textbf{100.0\%} & \textbf{90.0\%} & \textbf{85.0\%} \\

\bottomrule
\end{tabular}

\caption{\textbf{Evaluation Results on \textit{PnP} task.}
\mate{} achieves the best overall performance across the three stages of the \textit{PnP} task.
The cell highlighted in gray indicates high-frequency inference.}
\label{tab:pnp_results}
\vspace{-2.0em}
\end{table}

% \textbf{\mate{} effectively composes visual policies with complementary representations and viewpoints (Q3).} 
\textbf{\mate{} composes heterogeneous policies across representations, viewpoints, and inference rates (Q3).}
We further evaluate \mate{} in a heterogeneous visual-policy setting, where the two component policies differ in representation, viewpoint, and inference rate.
On the \textit{PnP} task shown in Fig.~\ref{fig:task-pnp}, the global-camera 3D policy provides global geometric guidance but often loses sight of the toy because of occlusion by the box walls. 
Consequently, RISE can approach the box yet often fails to reliably localize and grasp the toy. 
The wrist-camera 2D policy provides complementary close-range observations after the end effector enters the box. 
\mate{} combines global 3D guidance with local 2D observations to support occlusion-sensitive manipulation. 
As shown in Table~\ref{tab:pnp_results}, this composition improves execution across the Reach, Pick, and Place stages, particularly after the robot enters the box.

\textbf{Asynchronous composition provides modest but consistent gains over alternative fusion strategies.} 
% All fusion variants substantially outperformed the standalone 3D policy, confirming that the primary benefit comes from incorporating the complementary wrist-camera view. 
\mate{} achieved the best results across all three stages, although its improvement over the fusion baselines was relatively small. 
This suggests that increasing the inference frequency is less critical in this task than effectively combining complementary policy predictions.
Nevertheless, \mate{} preserves the faster policy's inference rate without sacrificing slower global guidance, demonstrating that the framework is not tied to a particular sensing modality or policy representation.

\subsection{Effect of Policy Inference Frequency}
\label{sec:exp-freq}

\textbf{Preserving higher policy inference rates improves composed-policy performance (Q4).}
We evaluate \mate{} under multiple force policy inference rates on the \textit{Flip} and \textit{Polish Curve} tasks.
Across these settings, the vision policy maintains a fixed, lower inference rate, while the force policy inference rate is varied.
As shown in Fig.~\ref{fig:freq}, increasing the force policy inference rate improves final task success on both tasks.
This trend suggests that more frequent incorporation of updated contact feedback contributes to the performance gains while complementing the lower-rate global visual guidance.
In both tasks, longer intervals between policy updates may allow contact errors to accumulate before corrective actions are generated.
A faster force policy can shorten this feedback-to-correction interval, while the slower vision policy continues to provide global task guidance.
These results indicate that synchronizing both policies at the slower inference rate limits the contribution of the force policy.
Within the evaluated range, preserving heterogeneous inference rates improves the use of complementary policies, supporting the rate-preserving design of \mate{}.

\section{Conclusion}
\label{sec:conclusion}
We presented \mate{}, a modular framework for asynchronous multi-rate diffusion policy composition.
\mate{} allows independently trained component policies with heterogeneous observation representations and inference rates to contribute guidance without being synchronized to the slowest component.
To enable consistent asynchronous composition, we introduced temporal overlap selection and derived a reference-frame rebasing rule that aligns diffusion variables generated at different robot states before weighted fusion.
We instantiated \mate{} by composing a lower-rate vision policy with a higher-rate force policy on two contact-rich manipulation tasks, and further demonstrated its generality through the composition of complementary 3D and 2D visual policies on an occluded pick-and-place task.
Across these real-world tasks, \mate{} outperforms synchronous composition, existing vision-only and multimodal policies, and a matched-frequency end-to-end baseline.
Our frequency analysis confirms that preserving native inference rates benefits composed policies.
Overall, \mate{} provides a modular framework for composing diffusion policies at heterogeneous inference rates across different modalities, representations, and viewpoints.

% We presented \mate{}, a latency-aware guidance fusion framework for asynchronous multi-rate diffusion policy composition.
% Unlike synchronized composition methods that require all modalities to be evaluated together, \mate{} allows modality-specific policies to operate at their native inference rates and contribute denoising guidance whenever available.
% To make asynchronous composition consistent, we derived a reference-frame rebasing rule for diffusion variables under relative action representations, enabling delayed guidance to be aligned before fusion.
% We instantiated \mate{} in contact-rich manipulation by composing a low-frequency vision policy with a high-frequency force policy.
% Experiments on two contact-rich tasks show that \mate{} improves the performance over synchronized fusion and slow-fast baselines.
% Our frequency analysis further confirms that preserving native inference rates benefits both standalone and composed policies.
% Overall, \mate{} provides a modular and practical interface for exploiting heterogeneous sensory modalities in diffusion-policy-based robot control.

\textbf{Limitations and Future Work.}
Although \mate{} improves over existing baselines, there is still room to further enhance task success and control stability.
Our current framework is developed for diffusion-based policies, extending it to flow matching~\cite{pi0,pi05} or other generative action modeling frameworks remains future work.
Moreover, our experiments focus on parallel-gripper manipulation.
Future work could apply \mate{} to dexterous hands~\cite{he2026scaling, trex}, where richer contact dynamics, higher-dimensional actions, and the need for high-frequency reactive control may further highlight the benefits of asynchronous multi-rate policy composition.

\printbibliography

@inproceedings{rdp,
  author    = {Xue, Han and Ren, Jieji and Chen, Wendi and Zhang, Gu and Fang, Yuan and Gu, Guoying and Xu, Huazhe and Lu, Cewu},
  title     = {Reactive Diffusion Policy: Slow-Fast Visual-Tactile Policy Learning for Contact-Rich Manipulation},
  booktitle = {RSS},
  year      = {2025}
}

@article{foar,
  author    = {He, Zihao and Fang, Hongjie and Chen, Jingjing and Fang, Hao-Shu and Lu, Cewu},
  title     = {FoAR: Force-Aware Reactive Policy for Contact-Rich Robotic Manipulation},
  journal   = {RA-L},
  year      = {2025}
}

@inproceedings{acp,
  author    = {Hou, Yifan and Liu, Zeyi and Chi, Cheng and Cousineau, Eric and Kuppuswamy, Naveen and Feng, Siyuan and Burchfiel, Benjamin and Song, Shuran},
  title     = {Adaptive Compliance Policy: Learning Approximate Compliance for Diffusion Guided Control},
  booktitle = {ICRA},
  year      = {2025}
}

@inproceedings{tavla,
  author    = {Zhang, Zongzheng and Xu, Haobo and Yang, Zhuo and Yue, Chenghao and Lin, Zehao and Gao, Huan-ang and Wang, Ziwei and Zhao, Hao},
  title     = {Elucidating the Design Space of Torque-aware Vision-Language-Action Models},
  booktitle = {CoRL},
  year      = {2025}
}

@inproceedings{forcevla,
  author    = {Yu, Jiawen and Liu, Hairuo and Yu, Qiaojun and Ren, Jieji and Hao, Ce and Ding, Haitong and Huang, Guangyu and Huang, Guofan and Song, Yan and Cai, Panpan and others},
  title     = {ForceVLA: Enhancing VLA Models with a Force-Aware MoE for Contact-Rich Manipulation},
  booktitle = {NeurIPS},
  year      = {2025}
}

@inproceedings{forcemimic,
  author    = {Liu, Wenhai and Wang, Junbo and Wang, Yiming and Wang, Weiming and Lu, Cewu},
  title     = {ForceMimic: Force-Centric Imitation Learning With Force-Motion Capture System for Contact-Rich Manipulation},
  booktitle = {ICRA},
  year      = {2025}
}

@inproceedings{maniwav,
  author    = {Liu, Zeyi and Chi, Cheng and Cousineau, Eric and Kuppuswamy, Naveen and Burchfiel, Benjamin and Song, Shuran},
  title     = {ManiWAV: Learning Robot Manipulation from In-the-Wild Audio-Visual Data},
  booktitle = {CoRL},
  year      = {2024}
}

@inproceedings{umi,
  author    = {Cheng Chi and Zhenjia Xu and Chuer Pan and Eric Cousineau and Benjamin Burchfiel and Siyuan Feng and Russ Tedrake and Shuran Song},
  title     = {Universal Manipulation Interface: In-The-Wild Robot Teaching Without In-The-Wild Robots},
  booktitle = {RSS},
  year      = {2024}
}

@inproceedings{poco,
  author    = {Lirui Wang and Jialiang Zhao and Yilun Du and Edward H. Adelson and Russ Tedrake},
  title     = {Policy Composition From and For Heterogeneous Robot Learning},
  booktitle = {RSS},
  year      = {2024}
}

@inproceedings{forcepolicy,
  author    = {Fang, Hongjie and Tang, Shirun and Mei, Mingyu and Qin, Haoxiang and He, Zihao and Chen, Jingjing and Feng, Ying and Wang, Chenxi and Liu, Wanxi and He, Zaixing and Lu, Cewu and Wang, Shiquan},
  title     = {Force Policy: Learning Hybrid Force-Position Control Policy under Interaction Frame for Contact-Rich Manipulation},
  booktitle = {RSS},
  year      = {2026}
}

@article{fdp,
  author    = {Liu, Chaoqi and Chen, Haonan and Høeg, Sigmund H. and Yao, Shaoxiong and Li, Yunzhu and Hauser, Kris and Du, Yilun},
  title     = {Flexible Multitask Learning With Factorized Diffusion Policy},
  journal   = {RA-L},
  year      = {2026}
}

@inproceedings{multimodal_consensus,
  author    = {Chen, Haonan and Xu, Jiaming and Chen, Hongyu and Hong, Kaiwen and Huang, Binghao and Liu, Chaoqi and Mao, Jiayuan and Li, Yunzhu and Du, Yilun and Driggs-Campbell, Katherine},
  title     = {Multi-Modal Manipulation via Multi-Modal Policy Consensus},
  booktitle = {ICRA},
  year      = {2026}
}

@inproceedings{crdagger,
  author    = {Xiaomeng Xu and Yifan Hou and Zeyi Liu and Shuran Song},
  title     = {Compliant Residual {DA}gger: Improving Real-World Contact-Rich Manipulation with Human Corrections},
  booktitle = {NeurIPS},
  year      = {2025}
}

@inproceedings{seehearfeel,
  author    = {Hao Li and Yizhi Zhang and Junzhe Zhu and Shaoxiong Wang and Michelle A. Lee and Huazhe Xu and Edward H. Adelson and Li Fei{-}Fei and Ruohan Gao and Jiajun Wu},
  title     = {See, Hear, and Feel: Smart Sensory Fusion for Robotic Manipulation},
  booktitle = {CoRL},
  year      = {2022}
}

@inproceedings{pi0,
  author    = {Black, Kevin and Brown, Noah and Driess, Danny and Esmail, Adnan and Equi, Michael and Finn, Chelsea and Fusai, Niccolo and Groom, Lachy and Hausman, Karol and Ichter, Brian and others},
  title     = {$\pi_0$: A Vision-Language-Action Flow Model for General Robot Control},
  booktitle = {RSS},
  year      = {2025}
}

@inproceedings{pi05,
  author    = {Black, Kevin and Brown, Noah and Darpinian, James and Dhabalia, Karan and Driess, Danny and Esmail, Adnan and Equi, Michael Robert and Finn, Chelsea and Fusai, Niccolo and Galliker, Manuel Y and others},
  title     = {$\pi_{0.5}$: a Vision-Language-Action Model with Open-World Generalization},
  booktitle = {CoRL},
  year      = {2025}
}

@inproceedings{gpc,
  author    = {Jiahang Cao and Yize Huang and Hanzhong Guo and Rui Zhang and Mu Nan and Weijian Mai and Jiaxu Wang and Hao Cheng and Jingkai Sun and Gang Han and Wen Zhao and Qiang Zhang and Yijie Guo and Qihao Zheng and Chunfeng Song and Xiao Li and Ping Luo and Andrew F. Luo},
  title     = {Compose Your Policies! Improving Diffusion-based or Flow-based Robot Policies via Test-time Distribution-level Composition},
  booktitle = {ICLR},
  year      = {2026}
}

@inproceedings{rise,
  author    = {Chenxi Wang and Hongjie Fang and Hao-Shu Fang and Cewu Lu},
  title     = {RISE: 3D Perception Makes Real-World Robot Imitation Simple and Effective},
  booktitle = {IROS},
  year      = {2024}
}

@inproceedings{diffusionpolicy,
  author    = {Cheng Chi and Siyuan Feng and Yilun Du and Zhenjia Xu and Eric Cousineau and Benjamin Burchfiel and Shuran Song},
  title     = {Diffusion Policy: Visuomotor Policy Learning via Action Diffusion},
  booktitle = {RSS},
  year      = {2023}
}

@inproceedings{act,
  author    = {Tony Z. Zhao and Vikash Kumar and Sergey Levine and Chelsea Finn},
  title     = {Learning Fine-Grained Bimanual Manipulation with Low-Cost Hardware},
  booktitle = {RSS},
  year      = {2023}
}

@inproceedings{images_that_sounds,
  author    = {Chen, Ziyang and Geng, Daniel and Owens, Andrew},
  title     = {Images that Sound: Composing Images and Sounds on a Single Canvas},
  booktitle = {NeurIPS},
  year      = {2024}
}

@inproceedings{compositional_ebm,
  author    = {Du, Yilun and Li, Shuang and Mordatch, Igor},
  title     = {Compositional Visual Generation and Inference with Energy Based Models},
  booktitle = {NeurIPS},
  year      = {2020}
}

@inproceedings{compositional_visual_generation,
  author    = {Nan Liu and Shuang Li and Yilun Du and Antonio Torralba and Joshua B. Tenenbaum},
  title     = {Compositional Visual Generation with Composable Diffusion Models},
  booktitle = {ECCV},
  year      = {2022}
}

@inproceedings{visualanagrams,
  author    = {Geng, Daniel and Park, Inbum and Owens, Andrew},
  title     = {Visual Anagrams: Generating Multi-View Optical Illusions with Diffusion Models},
  booktitle = {CVPR},
  year      = {2024}
}

@inproceedings{ddpm,
  author    = {Ho, Jonathan and Jain, Ajay and Abbeel, Pieter},
  title     = {Denoising Diffusion Probabilistic Models},
  booktitle = {NeurIPS},
  year      = {2020}
}

@inproceedings{ddim,
  author    = {Jiaming Song and Chenlin Meng and Stefano Ermon},
  title     = {Denoising Diffusion Implicit Models},
  booktitle = {ICLR},
  year      = {2021}
}

@inproceedings{rotation_6d,
  author    = {Zhou, Yi and Barnes, Connelly and Lu, Jingwan and Yang, Jimei and Li, Hao},
  title     = {On the Continuity of Rotation Representations in Neural Networks},
  booktitle = {CVPR},
  year      = {2019}
}

@inproceedings{generative_skill_chaining,
  author    = {Mishra, Utkarsh A. and Xue, Shangjie and Chen, Yongxin and Xu, Danfei},
  title     = {Generative Skill Chaining: Long-Horizon Skill Planning with Diffusion Models},
  booktitle = {CoRL},
  year      = {2023}
}

@inproceedings{cdgs,
  author    = {Mishra, Utkarsh Aashu and He, David and Chen, Yongxin and Xu, Danfei},
  title     = {Compositional Diffusion with Guided Search for Long-Horizon Planning},
  booktitle = {ICLR},
  year      = {2026}
}

@inproceedings{ebm_zero_shot_planner,
  author    = {Du, Yilun and Yang, Mengjiao and Florence, Pete and Xia, Fei and Wahid, Ayzaan and Ichter, Brian and Sermanet, Pierre and Yu, Tianhe and Abbeel, Pieter and Tenenbaum, Joshua B. and Lynch, Corey and Tompson, Jonathan},
  title     = {Energy-Based Models are Zero-Shot Planners for Compositional Scene Rearrangement},
  booktitle = {RSS},
  year      = {2023}
}

@inproceedings{cage,
  author    = {Xia, Shangning and Fang, Hongjie and Fang, Hao-Shu and Lu, Cewu},
  title     = {CAGE: Causal Attention Enables Data-Efficient Generalizable Robotic Manipulation},
  booktitle = {ICRA},
  year      = {2025}
}

@inproceedings{geng2024factorized,
  author    = {Geng, Daniel and Park, Inbum and Owens, Andrew},
  title     = {Factorized Diffusion: Perceptual Illusions by Noise Decomposition},
  booktitle = {ECCV},
  year      = {2024}
}

@inproceedings{dexperts,
  author    = {Liu, Alisa and Sap, Maarten and Lu, Ximing and Swayamdipta, Swabha and Bhagavatula, Chandra and Smith, Noah A. and Choi, Yejin},
  title     = {DExperts: Decoding-Time Controlled Text Generation with Experts and Anti-Experts},
  booktitle = {ACL},
  year      = {2021}
}

@inproceedings{fudge,
  author    = {Yang, Kevin and Klein, Dan},
  title     = {FUDGE: Controlled Text Generation With Future Discriminators},
  booktitle = {NAACL},
  year      = {2021}
}

@inproceedings{gedi,
  author    = {Krause, Ben and Gotmare, Akhilesh Deepak and McCann, Bryan and Keskar, Nitish Shirish and Joty, Shafiq and Socher, Richard and Rajani, Nazneen Fatema},
  title     = {GeDi: Generative Discriminator Guided Sequence Generation},
  booktitle = {Findings of EMNLP},
  year      = {2021}
}

@inproceedings{neurologic,
  author    = {Lu, Ximing and West, Peter and Zellers, Rowan and Le Bras, Ronan and Bhagavatula, Chandra and Choi, Yejin},
  title     = {NeuroLogic Decoding: (Un)supervised Neural Text Generation with Predicate Logic Constraints},
  booktitle = {NAACL},
  year      = {2021}
}

@misc{tdk,
  author    = {Flexiv Ltd},
  title     = {Flexiv Teleoperation Development Kit (TDK)},
  note      = {Accessed January 2026}
}

@inproceedings{prasad2024consistency,
  author    = {Prasad, Aaditya and Lin, Kevin and Wu, Jimmy and Zhou, Linqi and Bohg, Jeannette},
  title     = {Consistency Policy: Accelerated Visuomotor Policies via Consistency Distillation},
  booktitle = {RSS},
  year      = {2024}
}

@inproceedings{lu2024manicm,
  author    = {Zifeng Gao and Guanxing Lu and Tianxing Chen and Wenxun Dai and Ziwei Wang and Chao Shang and Wenbo Ding and Yansong Tang},
  title     = {ManiCM: Real-time 3D Diffusion Policy via Consistency Model for Robotic Manipulation},
  booktitle = {RCAR},
  year      = {2026}
}

@inproceedings{rtc,
  author    = {Black, Kevin and Galliker, Manuel Y. and Levine, Sergey},
  title     = {Real-Time Execution of Action Chunking Flow Policies},
  booktitle = {NeurIPS},
  year      = {2025}
}

@inproceedings{hoeg2025streaming,
  author    = {H{\o}eg, Sigmund H. and Du, Yilun and Egeland, Olav},
  title     = {Streaming Diffusion Policy: Fast Policy Synthesis with Variable Noise Diffusion Models},
  booktitle = {ICRA},
  year      = {2025}
}

@inproceedings{duan2025realtime,
  author    = {Duan, Yufei and Yin, Hang and Kragic, Danica},
  title     = {Real-Time Iteration Scheme for Diffusion Policy},
  booktitle = {IROS},
  year      = {2025}
}

@inproceedings{hirt,
  title     = {HiRT: Enhancing Robotic Control with Hierarchical Robot Transformers},
  author    = {Zhang, Jianke and Guo, Yanjiang and Chen, Xiaoyu and Wang, Yen-Jen and Hu, Yucheng and Shi, Chengming and Chen, Jianyu},
  booktitle = {CoRL},
  year      = {2024}
}

@inproceedings{fastinslow,
  title     = {Fast-in-Slow: A Dual-System VLA Model Unifying Fast Manipulation within Slow Reasoning},
  author    = {Chen, Hao and Liu, Jiaming and Gu, Chenyang and Liu, Zhuoyang and Zhang, Renrui and Li, Xiaoqi and He, Xiao and Guo, Yandong and Fu, Chi-Wing and Zhang, Shanghang and Heng, Pheng-Ann},
  booktitle = {NeurIPS},
  year      = {2025}
}

@article{trex,
  title   = {T-Rex: Tactile-Reactive Dexterous Manipulation},
  author  = {Niu, Dantong and Liu, Zhuoyang and Wang, Zekai and Shao, Boning and Yin, Zhao-Heng and Pai, Anirudh and Sharma, Yuvan and Saravalle, Stefano and Zheng, Ruijie and Wang, Jing and Punamiya, Ryan and Xu, Mengda and Xie, Yuqi and Jiang, Yunfan and Fu, Letian and Kallidromitis, Konstantinos and Gioia, Matteo and Zhang, Junyi and Ge, Jiaxin and Feng, Haiwen and Galasso, Fabio and Zhan, Wei and Chan, David M. and Bai, Yutong and Herzig, Roei and Lei, Jiahui and Li, Fei-Fei and Goldberg, Ken and Malik, Jitendra and Abbeel, Pieter and Zhu, Yuke and Xu, Danfei and Fan, Jim and Darrell, Trevor},
  journal = {arXiv preprint arXiv:2606.17055},
  year    = {2026}
}

@inproceedings{cpm,
  title     = {Composable Part-Based Manipulation},
  author    = {Liu, Weiyu and Mao, Jiayuan and Hsu, Joy and Hermans, Tucker and Garg, Animesh and Wu, Jiajun},
  booktitle = {CoRL},
  year      = {2023}
}

@inproceedings{he2026scaling,
  title     = {Scaling Cross-Embodiment World Models for Dexterous Manipulation},
  author    = {He, Zihao and Ai, Bo and Mu, Tongzhou and Liu, Yulin and Wan, Weikang and Fu, Jiawei and Du, Yilun and Christensen, Henrik I. and Su, Hao},
  booktitle = {IROS},
  year      = {2026}
}

\clearpage
\section*{APPENDIX}

\section{Method Implementation Details}
\label{app:method}

\subsection{Visual Policy Details}
\label{app:visual_policy}

% Our visual policy follows the original RISE policy design~\cite{rise}.
% We keep the visual input representation, network architecture, diffusion action prediction head, and training procedure unchanged.
% The main difference lies in the action coordinate representation.
% The original RISE policy represents predicted actions in the camera global coordinate frame, whereas our implementation represents actions in a relative coordinate frame with respect to the current end-effector pose.

% This relative action representation is introduced to support asynchronous policy composition.
% Since the visual policy and the force/torque policy may produce guidance at different timestamps, their predicted action trajectories must be expressed in a form that can be transformed across reference frames.
% By representing visual actions relative to the current end-effector pose, we can rebase delayed visual predictions before composing them with high-frequency force/torque guidance.
% All other visual policy hyperparameters and implementation choices follow RISE~\cite{rise}.

Our 3D visual policy follows the original RISE design~\cite{rise}.
We retain its point-cloud input representation, network architecture, diffusion action prediction head, and training procedure.
The main modification is the action coordinate representation: instead of predicting actions in the camera coordinate frame, we represent actions relative to the end-effector pose at the policy query time.
This representation allows predictions generated at different query times to be rebased into a common action reference frame before asynchronous composition.
All other hyperparameters and implementation choices of the 3D visual policy follow RISE.

For the \textit{Occluded Pick-and-Place} task, we additionally train a standard RGB-conditioned Diffusion Policy using observations from the wrist camera.
The policy takes a single RGB image at a resolution of $720 \times 1280$ and encodes it into a feature using ResNet-18.
Both vision policies predict a 100-step action chunk at 50\,Hz, corresponding to a 2\,s prediction horizon.
Each action uses a relative representation with 10 dimensions: 3D translation, a column-wise 6D rotation representation, and a gripper command.

\subsection{Force Policy Details}
\label{app:ft_policy}

We train a force-conditioned diffusion policy as the low-dimensional force policy.
In the main compositional policy, the force policy takes a 1s force/torque history as input, corresponding to 100 steps at 100Hz with six force/torque channels, and predicts a 1s future action chunk, corresponding to 50 action steps at 50Hz.
Each action is represented in a 10D space, including 3D translation, 6D rotation, and one gripper command.
Rotations are parameterized by the column-wise 6D rotation representation.
The force/torque stream is expressed in the TCP frame and normalized to $[-1, 1]$ using fixed per-axis force/torque bounds.
Actions are represented in a relative formulation, using the current end-effector pose as the base pose. The key hyperparameters of the force policy are summarized in Table~\ref{tab:ft_policy_hyperparams}.

For the standalone force policy analysis in Appendix~\ref{app:additional_results}, we additionally vary the force/torque input frequency and inference frequency to study their effects on performance.
% Unless otherwise specified, the hyperparameters in Table~\ref{tab:ft_policy_hyperparams} correspond to the force policy used in our main compositional policy.

\begin{table}[h]
\centering
\small
\begin{tabular}{lc}
\toprule
\textbf{Item} & \textbf{Setting} \\
\midrule
Force input length & 1s \\
Force input frequency & 100Hz \\
Force input steps & 100 \\
Force/torque channels & 6 \\
Force feature dimension & 16 \\
Force encoder & MLP: $600 \rightarrow 256 \rightarrow 16$ \\
Action chunk length & 50 \\
Action chunk frequency & 50Hz \\
Action representation & 10D: 3D translation, 6D rotation, gripper \\
U-Net channels & $[32, 64]$ \\
DDPM training steps & 100 \\
DDIM inference steps & 10 \\
Optimizer & AdamW \\
Learning rate & $3 \times 10^{-4}$ \\
Batch size & 256 \\
Epochs & 100 \\
\bottomrule
\end{tabular}\vspace{0.2cm}
\caption{
\textbf{Key Hyperparameters of the Force Policy.}
The table reports the default configuration with a 100 Hz force/torque input frequency. In the standalone frequency study, the input dimension of the force encoder is adjusted according to the input frequency, while the remaining architecture and training settings are kept consistent unless otherwise specified.
}
\label{tab:ft_policy_hyperparams}
\end{table}

\subsection{Inference and Asynchronous Composition Details}

% weights
\textbf{Horizon-Dependent Composition Weights.}
During asynchronous composition, the vision and force/torque policies provide guidance at different update rates. The vision policy is queried less frequently and provides global, long-horizon guidance, while the force/torque policy is queried more frequently and provides fresher local guidance for contact-sensitive correction. Therefore, when fusing the two guidance signals, we use a latency-aware weighting schedule that changes along the selected action horizon.

Specifically, for each action index within the force policy horizon, we compute the vision weight by linearly decreasing it from a high value to a low value. 
Let $i$ denote the action index and $I_{\max}$ denote the maximum index of the selected horizon. 
We first compute a normalized horizon ratio $r_i = {i}/{I_{\max}}.$    
The vision weight is then computed by linear interpolation:
\begin{equation}
    w_i^V = w_{\mathrm{start}}^V 
    + r_i \left( w_{\mathrm{end}}^V - w_{\mathrm{start}}^V \right).
\end{equation}
In our implementation for vision--force composition, $w_{\mathrm{start}}^V = 0.8$, $w_{\mathrm{end}}^V = 0.2$, and $I_{\max}=50$. 
Therefore, the vision weight starts from 0.8 at the beginning of the selected horizon and gradually decreases to 0.2 by the end of the horizon. 
The force weight is defined as the complementary value $w_i^F = 1 - w_i^V$.
This creates an index-dependent weighting schedule: early actions rely more on vision guidance, while later actions rely more on force/torque guidance.

% This design reflects the temporal reliability of the two modalities. Near the beginning of the selected horizon, the delayed vision guidance is still relatively close to the current robot state and provides useful global task direction. Therefore, the fused guidance relies more on the vision policy. As the horizon progresses, the visual prediction becomes less temporally fresh, while the force/torque policy provides more up-to-date local contact feedback. The fusion therefore gradually shifts more weight to the force/torque guidance.

% runtime analysis
\textbf{Runtime Analysis.}
We report the runtime of the proposed asynchronous vision--force composition during evaluation. 
With our implementation, the mean policy inference time after composition is approximately 40ms, which enables 25Hz policy inference. 
In contrast to synchronous composition, where all modalities are typically constrained by the slowest policy, our asynchronous formulation allows the low-dimensional force branch to be updated at a higher frequency while still incorporating visual guidance when available.

\label{app:experiments}

\section{Experiment Details}

\subsection{Setup}

\begin{figure}[t]
    \centering
    \includegraphics[width=\columnwidth]{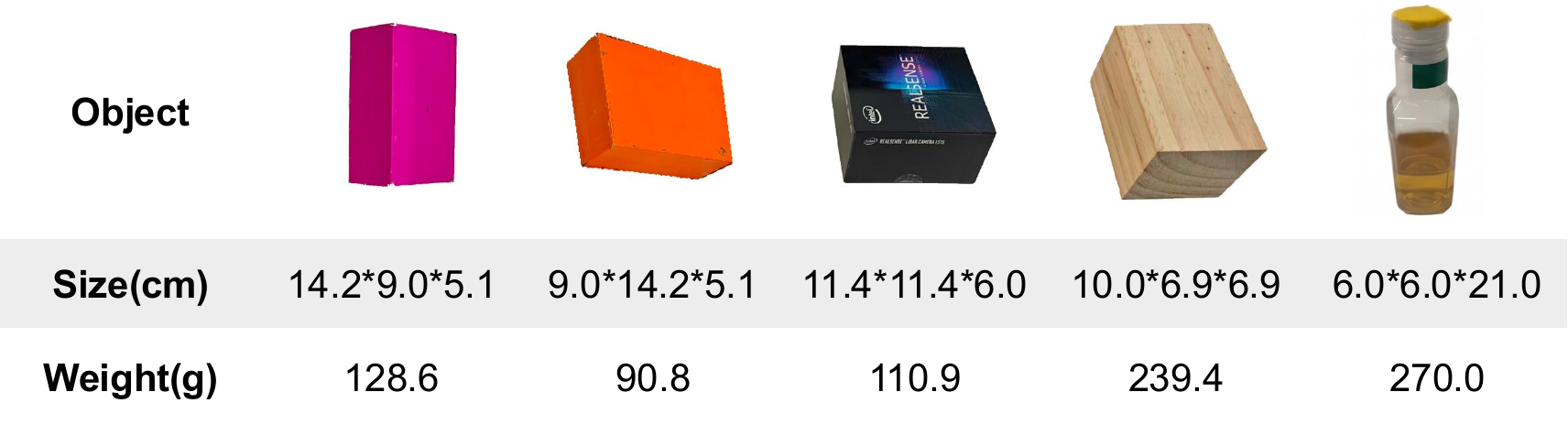}
    \caption{\textbf{Evaluation Objects for the Low-dimensional Force Policy on \textit{Flip}.}
    The policy is tested on five objects with varying sizes and weights. Only the first purple box is seen during training, while the other four objects are unseen.}
    \label{fig:objects}
\end{figure}

\begin{figure*}[t]
    \centering
    \includegraphics[width=\linewidth]{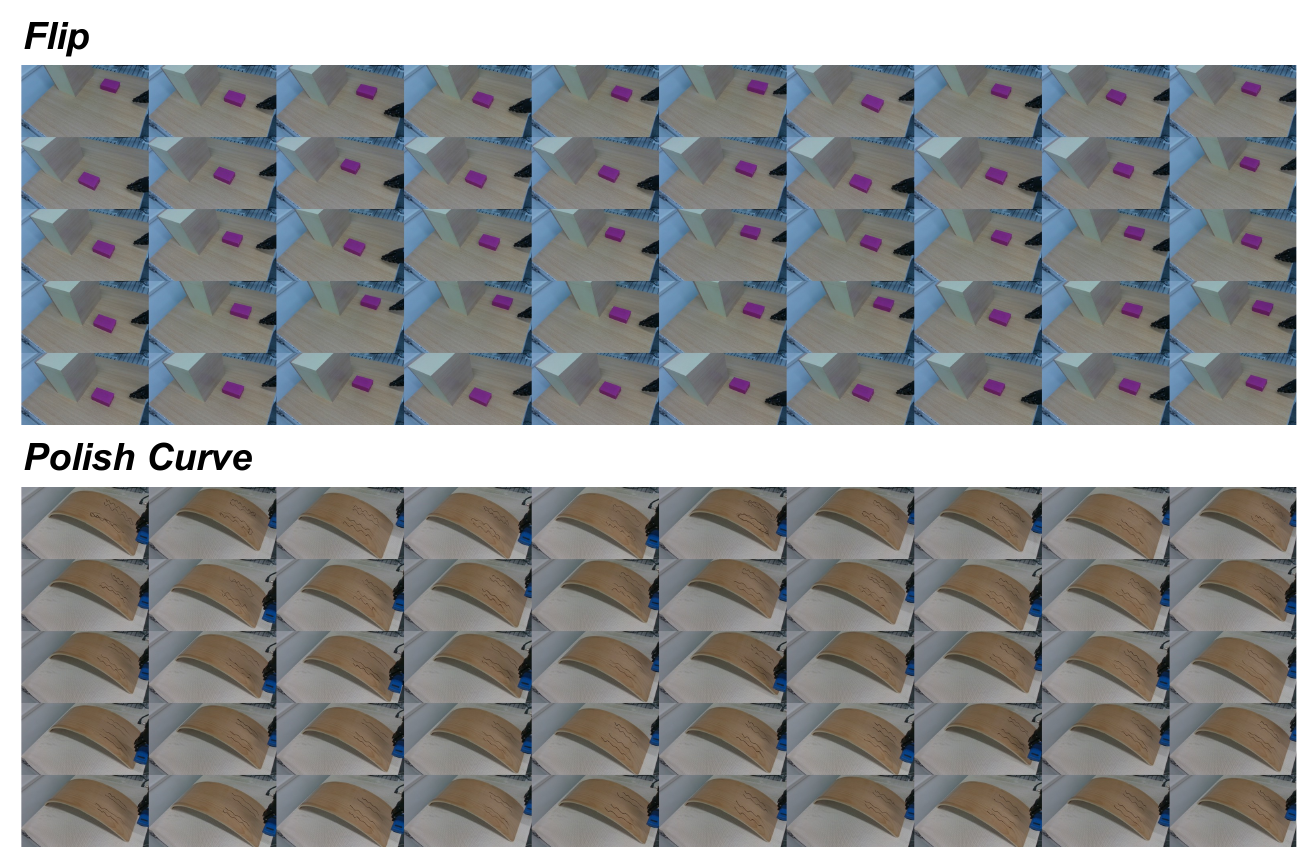}
    \caption{\textbf{Overview of Demonstrations for each Tasks.} We visualize the initial configurations from 50 demonstrations for each task.
    }
    \label{fig:dataset_demonstration}
\end{figure*}

\textbf{Data Collection.} 
We visualize the initial configurations of all 50 demonstrations for each task in Fig.~\ref{fig:dataset_demonstration}. The objects are randomly placed within the workspace during demonstration collection.

\begin{figure*}[t]
    \centering
    \includegraphics[width=\linewidth]{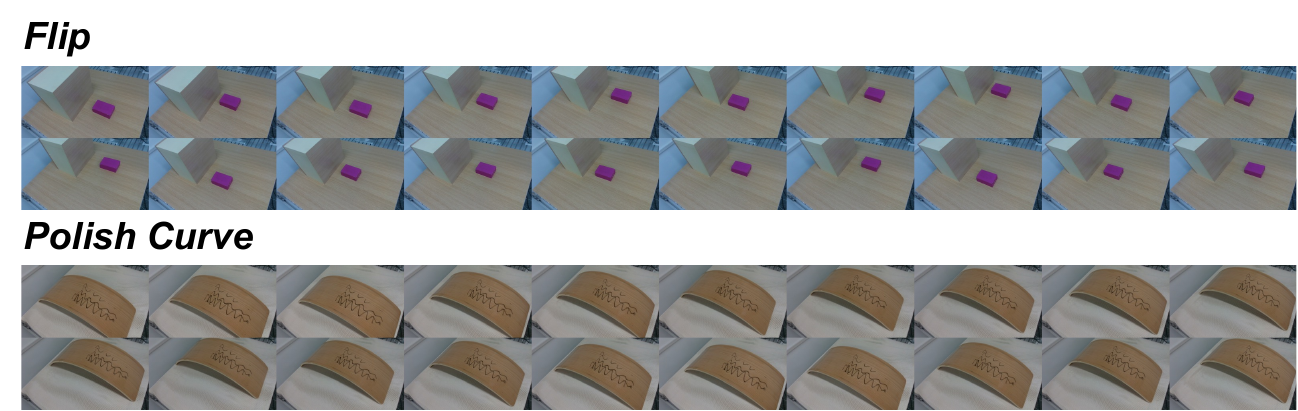}
    \caption{\textbf{Evaluation Initial Configurations.}
    For each task, we evaluate each policy on 20 different initial configurations. 
    The object is randomly placed at the beginning of each trial, and these evaluation configurations are different from those used in the training demonstrations.}
    \label{fig:eval_init_config}
\end{figure*}

\textbf{Evaluation Protocols.}
To ensure a fair and reproducible comparison, we adopt an identical evaluation protocol across all policies. As shown in Fig.~\ref{fig:eval_init_config}, we pre-sample a fixed set of initial object placements that are uniformly distributed over the workspace, and reuse exactly the same set of placements for every policy. 
This isolates the effect of the policy itself from variation in the test layout, so that all methods are evaluated under identical initial conditions. 
Unless otherwise stated, the main results in Table~\ref{tab:flip_polish_curve_results} use the seen object and randomize only its placement. Each policy is evaluated over 20 consecutive trials per task, and success rates are computed accordingly.

\subsection{Additional Experiments Results}
\label{app:additional_results}

\begin{figure*}[t]
    \centering
    \includegraphics[width=\linewidth]{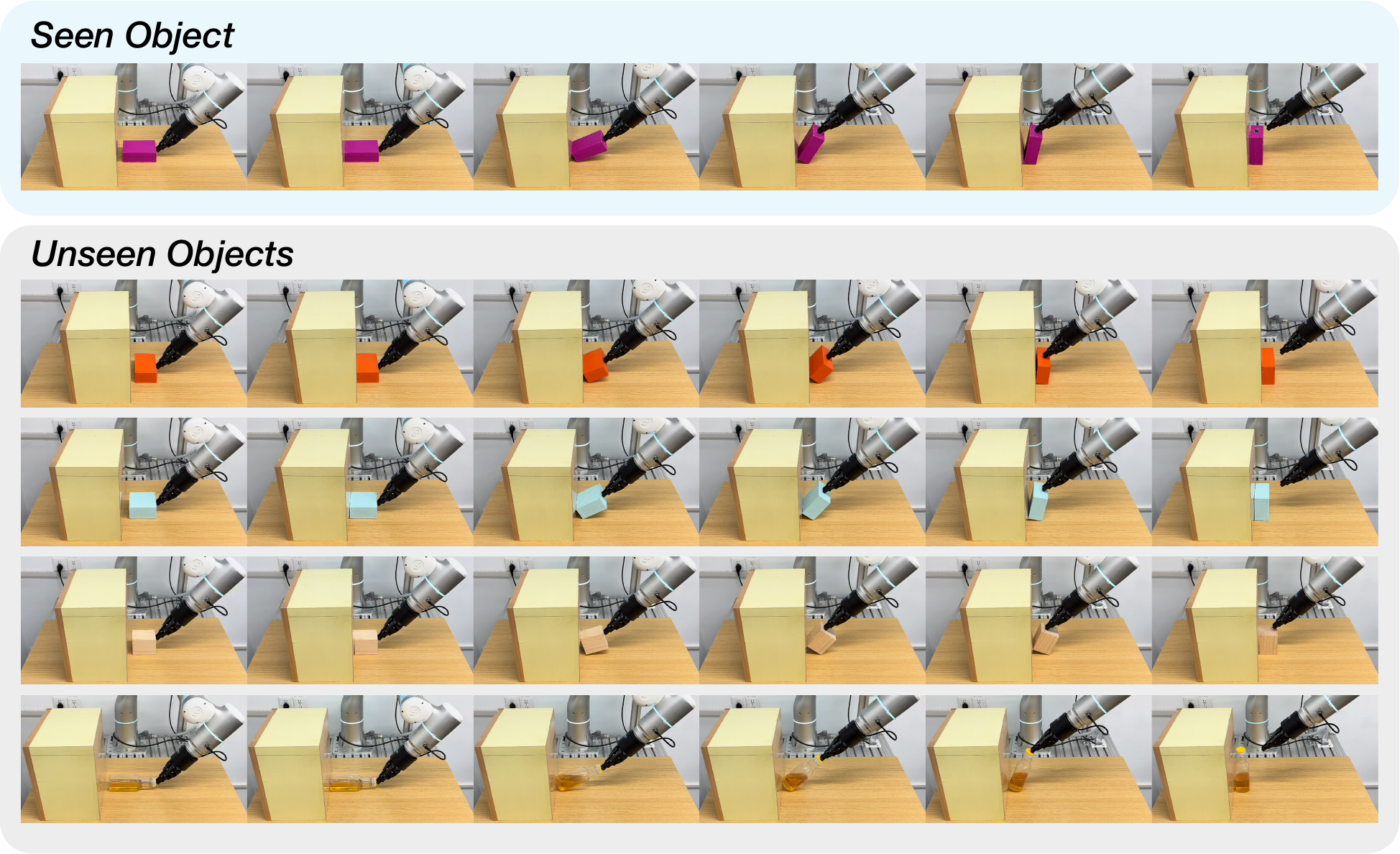}
    \caption{
    \textbf{Evaluation Setup for the Standalone Low-dimensional Policy.}
    We evaluate the standalone low-dimensional force/torque policy on five objects with different shapes, materials, and contact properties. 
    The top row shows the \textit{Purple Box} used in training, while the bottom rows show additional test objects used during evaluation. 
    This setup is used to compare the standalone policy under different input-frequency and inference-frequency configurations.
    }
    \label{fig:lowdim_policy_scene}
\end{figure*}

% lowdim force policy
\begin{figure*}[t]
    \centering
    \includegraphics[width=\linewidth]{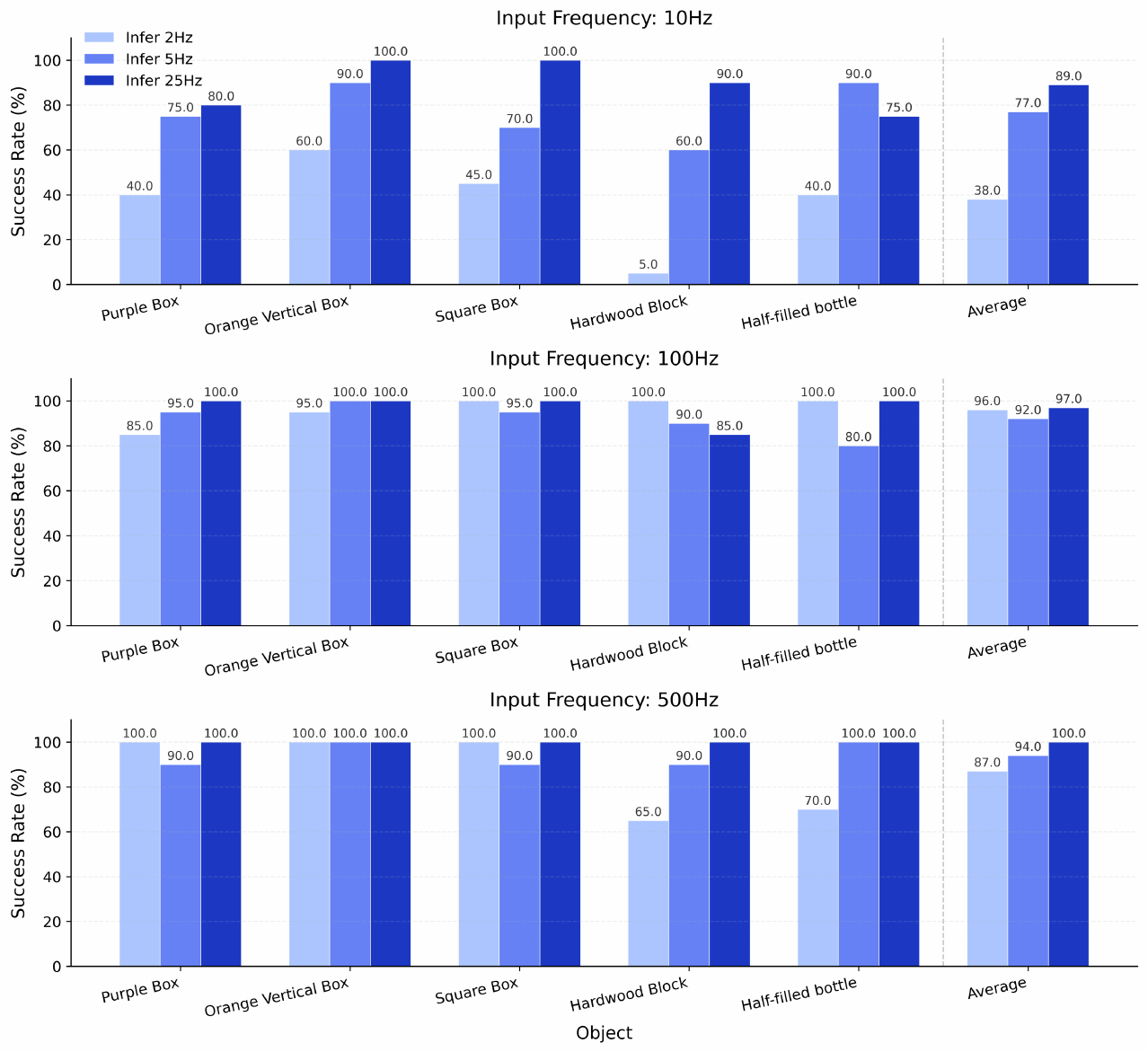}
    \caption{\textbf{Detailed Standalone Force Policy Results under Different Input and Inference Frequencies.}
    We evaluate force/torque policies trained with different input frequencies, including 10Hz, 100Hz, and 500Hz, and test each policy under different inference frequencies, including 2Hz, 5Hz, and 25Hz. 
    Results are reported across five test objects, with the average success rate shown on the right of each subplot. 
    Increasing the inference frequency generally improves success rate, especially under low-frequency input, while high-frequency input combined with high-frequency inference achieves the best overall performance.}
    \label{fig:force_policy_freq_detail}
\end{figure*}

We additionally evaluate the standalone low-dimensional force/torque policy to analyze the effect of input and inference frequencies, as motivated in the teaser and Introduction.
Unlike the full-task evaluation in Table~\ref{tab:flip_polish_curve_results}, this experiment initializes the robot near the object in a pre-contact configuration, isolating the contact-sensitive flipping stage to study the impact of inference frequency.
Fig.~\ref{fig:lowdim_policy_scene} shows the evaluation setup.
The policy is tested on five objects, including the \textit{Purple Box} used in training and four unseen objects with different shapes, materials, and contact properties.
This controlled setup allows us to systematically evaluate how input and inference frequencies affect contact-sensitive manipulation.

We further report the detailed success rates of the standalone force policy under different input-frequency and inference-frequency configurations in Fig.~\ref{fig:force_policy_freq_detail}. 
For each input frequency, we keep the force history window fixed to 1s and vary the number of input steps according to the input frequency. 
We evaluate each policy with three inference frequencies, 2Hz, 5Hz, and 25Hz, across 5 objects. 
Each input-frequency and inference-frequency configuration is evaluated with 10 trials per object. 
The results show that increasing the inference frequency generally improves the success rate, especially when the input frequency is low.
These results support our motivation that force feedback should not be bottlenecked by a low inference rate, and that high-frequency local feedback is important for contact-sensitive manipulation.

\subsection{Failure Analysis}
% LAG-Fusion: 1. modifying RISE to relative action representation makes it bad.

We further analyze representative failure modes for the \textit{Flip} and \textit{Polish Curve} tasks. 
Overall, the failures are closely related to insufficient reactivity to force feedback and the difficulty of balancing global visual guidance with local contact feedback.

\textbf{Flip.}
For \textit{Flip}, we observe three common failure modes in the baselines. 
First, for force based baselines, the force signal can be noisy during the contact-free pushing stage. 
This noise may cause the robot arm to drift from the desired contact position with the small box, making the subsequent flipping motion difficult to complete. 
Second, during the flipping stage, some baselines do not run at a sufficiently high inference frequency and therefore cannot react quickly to force feedback. 
As a result, the robot fails to adjust its end-effector position in time after contact, leading to missed or incomplete flips. 
Third, vision-only policies do not have access to force feedback and may prematurely switch to the flipping motion before the small box establishes contact with the target box.

\textbf{Polish Curve.}
For \textit{Polish Curve}, the main challenge is to maintain stable and sufficiently strong contact while following the curved trajectory. 
Baselines with low inference frequency often fail to consistently maintain large contact force, resulting in weak or unstable polishing behavior. 
Vision-only policies may also enter the polishing stage before actual contact is established, since they cannot directly sense the contact state. 

Compared with these baselines, \mate{} better balances visual and force guidance. 
The visual policy provides global guidance, while the force/torque policy provides local guidance for contact-sensitive correction. 
And it allows each modalities run at its native frequency. 
This makes the composed policy more reactive during contact transitions and more stable during sustained contact, leading to stronger performance on both tasks.

\textbf{Occluded Pick-and-Place (PnP)}. For \textit{PnP}, the main challenge arises from severe object occlusion caused by the high-walled box. The global-camera 3D policy provides useful guidance for reaching the box but often fails to localize the toy once the robot enters the box, resulting in unsuccessful grasps and subsequent placement failures. Although the wrist-camera 2D policy provides complementary close-range observations, effectively coordinating global spatial guidance with local visual feedback remains challenging. As a result, existing fusion baselines still exhibit failures during grasping and placement. \mate{} mitigates these limitations by asynchronously composing the two visual policies, preserving global guidance while incorporating more frequent local observations. Nevertheless, occasional failures remain during the picking and placing stages, indicating room for further improvement in occlusion-sensitive manipulation.

\end{document}